\documentclass[10pt]{article} 

\usepackage[dvips]{graphicx}
\usepackage[preprint]{tmlr}

\usepackage{amsmath,amsfonts,bm}

\newcommand{\Yh}{\hat Y}
\newcommand{\Yt}{\tilde Y}
\newcommand{\Cb}{\bar{\mathcal{C}}}
\newcommand{\Sb}{\bar S}
\newcommand{\Db}{\bar{\mathcal{D}}}
\newcommand{\MI}{\mathcal{M}_I}
\newcommand{\pbar}{\bar p}

\def\eqref#1{equation~\ref{#1}}

\def\1{\bm{1}}

\DeclareMathAlphabet{\mathsfit}{\encodingdefault}{\sfdefault}{m}{sl}
\SetMathAlphabet{\mathsfit}{bold}{\encodingdefault}{\sfdefault}{bx}{n}

\usepackage{amsthm}
\usepackage{dsfont}

\providecommand{\Yh}{\hat Y}
\providecommand{\Yt}{\tilde Y}
\providecommand{\MI}{\mathcal M_I}
\providecommand{\Cb}{\bar{\mathcal C}}
\providecommand{\Sb}{\bar S}
\providecommand{\Db}{\bar{\mathcal D}}
\providecommand{\pbar}{\bar p}

\makeatletter
\@ifundefined{proposition}{\newtheorem{proposition}{Proposition}}{}
\@ifundefined{definition}{\theoremstyle{definition}\newtheorem{definition}{Definition}\theoremstyle{plain}}{}
\@ifundefined{assumption}{\newtheorem{assumption}{Assumption}}{}
\@ifundefined{observation}{\newtheorem{observation}{Observation}}{}
\@ifundefined{remark}{\theoremstyle{remark}\newtheorem{remark}{Remark}\theoremstyle{plain}}{}
\makeatother

\usepackage{hyperref}
\usepackage{url}

\title{An Information Budget for Label-Noise Memorization, and What Statistical Complexity Tracks: A Study with Multilayer Perceptrons}

\author{\name Nancy Baygorrea C.\email nbaygorrea@gmail.com \\
      \addr Laboratory of Mathematical Sciences\\
      State University of Northern Rio de Janeiro
      \AND
      \name Otávio da Fonseca Martins Gomes \email ogomes@mn.ufrj.br\\
      \addr Centre for Mineral Technology \\
      Postgraduate Program in Geosciences,\\
      National Museum, Rio de Janeiro}

\def\month{MM}  
\def\year{YYYY} 
\def\openreview{\url{https://openreview.net/forum?id=XXXX}} 

\begin{document}

\maketitle

\begin{abstract}

How much of a noisy training set has a network memorized, and when? We study this question for a multilayer perceptron and ask what a statistical complexity (entropy-disequilibrium) description of its predictions adds. Standard information identities decompose the information that the predictions carry about the inputs into signal, memorization ($I(\Yh;\Yt \mid Y)$) and heterogeneity terms. The memorization term is zero in the population for predictors that do not depend on the realized class-conditional noise. Fano's inequality bounds it from below through training accuracies, but no upper bound follows from the memorized fraction $M$. In 89 runs on MNIST with ten pre-specified hypotheses (five supported, four not supported, one not testable under the protocol), the normalized memorization term rose with $M$ on the training set, stayed near zero on held-out samples with freshly drawn noisy labels, and was close to its Fano bound. Unlike $M$, it has no chance level by construction, although a chance correction of $M$ largely removed this difference. With every label replaced by an incorrect class, test accuracy fell to 0.5\% while the predictions retained information about the true class. The peak of mean statistical complexity preceded memorization at corruption rates up to 40\%; at higher rates it fell within the memorization phase, close to the epoch at which mean training confidence reached about 0.6, so the two could not be separated. As a stopping rule it matched established rules only at low noise, and it detected corrupted samples less well than simple label-free scores. An exploratory extension of 280 runs (Fashion-MNIST, pair-flip noise and four training variants) gave similar results under uniform noise but not under pair-flip noise, where the complexity peak occurred within the first two epochs. All conclusions are limited to this architecture, two small datasets and synthetic noise. 

\end{abstract}

\section{Introduction}
\label{sec:intro}

Deep neural networks have been shown to fit training sets with randomly assigned labels~\citep{Zhang2021}. When exposed to partial label corruption, training often exhibits an early-learning pattern: models tend to learn consistent semantic structure before increasingly fitting corrupted labels~\citep{Arpit2017,LiuELR2020,Song2020Survey}, a behavior that has also been analyzed theoretically for gradient descent with early stopping~\citep{LiSoltanolkotabiOymak2020}. Understanding this progression may help characterize generalization under label noise and inform early-stopping strategies~\citep{Bai2021,YuanLabelWave2024}. It also raises two research questions. \textbf{RQ1:} how much information about the label noise is reflected in a model's predictions, and how does it relate to accuracy-based measures of memorization? \textbf{RQ2:} what do observable summaries of those predictions, in particular statistical complexity, track during training, and are they useful as stopping rules or as detectors of corrupted samples?

Several operational perspectives on memorization address different aspects of these questions. Counterfactual approaches, motivated by the role of memorization in learning long-tailed distributions~\citep{Feldman2019}, assess memorization through the effect of including or removing individual examples from the training set~\citep{FeldmanZhang2020,ZhangCounterfactual2023}. Information-theoretic approaches examine the information that learned weights retain about label noise~\citep{Harutyunyan2020}, or bound generalization through the mutual information~\citep{XuRaginsky2017} or the conditional mutual information~\citep{Steinke2020} between the training data and the output of a learning algorithm. Representation-based work reports that networks that memorize shuffled training labels can retain, in their internal layers, information that supports generalization to the true labels~\citep{KethaRamaswamy2026}. On the practical side, corrupted samples are often identified from training dynamics, for example through small losses~\citep{Han2018,LiDivideMix2020}, margins~\citep{Pleiss2020}, forgetting events~\citep{Toneva2019} or early-training error norms~\citep{Paul2021}. Here, we work in function space, focusing on the classifier's conditional predictive distributions $p_\theta(\cdot\mid x)$. We use the dependence between predictions and observed labels, conditional on the true class, as an operational measure of label-noise memorization, and investigate how this quantity relates to accuracy-based measures of label-noise fitting and to geometric summaries of the predictions.

Our geometric description draws on a statistical-physics framework that characterizes probability distributions through normalized Shannon entropy $S$, disequilibrium $\mathcal{D}$ relative to the uniform distribution, and statistical complexity $\mathcal{C}=S \cdot \mathcal{D}$~\citep{LopezRuiz1995,Martin2006,Rosso2007}. These quantities map each predictive distribution to a point in an admissible region of the $(S,\mathcal{D})$ plane, allowing its evolution to be followed during training. Because complexity vanishes at both the uniform and deterministic limits, it characterizes intermediate features of predictive geometry rather than directly measuring memorization. The order of aggregation also matters: computing statistics after averaging predictions can obscure differences among individual outputs. Confident predictions distributed across classes, for example, can have a nearly uniform mean despite having low per-sample entropy.

We therefore consider the mean per-sample complexity $\Cb$ and the within-group variance $V$ of per-sample complexity, with groups defined by the noisy label, as complementary descriptions of training dynamics. The former summarizes the complexity of individual predictive distributions, while the latter measures their dispersion within groups that mix clean and corrupted samples. To provide an information-theoretic reference, we consider a fixed trained model and define $\hat{Y}$ as a categorical variable sampled from $p_\theta(\cdot\mid X)$. The entropy of the mean predictive distribution minus the mean per-sample entropy then equals $I(X;\hat{Y})$ and  with entropies normalized by $\ln N$ for $N$ classes, the difference equals $I(X;\hat{Y})/\ln N$. This construction is similar in form to entropy-difference criteria used in Bayesian active learning~\citep{Houlsby2011}, although the random variables and their interpretation differ. In the main grid of our experiments, the normalized entropy of the mean prediction remained above $0.99$, so marginal entropy alone offers limited separation between training regimes. Under the empirical distribution of the training set, the chain rule decomposes this predictive information exactly, without independence assumptions, into a signal term $I(\hat{Y};Y)$, a memorization term $I(\hat{Y};\tilde{Y}\mid Y)$ and a heterogeneity term $I(\hat{Y};X\mid Y, \tilde{Y})$, where $Y$ and $\tilde{Y}$ denote the clean and observed labels, respectively; the assumption of class-conditional noise is needed only to interpret the population value of the memorization term. This decomposition provides an information budget against which the geometric observables can be evaluated. Together with a Fano-type bound and an analysis of chance-level label matching, it helps distinguish the information associated with label-noise fitting from the predictive geometry summarized by statistical complexity.
\paragraph{Pre-specified expectations.} Before the main runs, we specified ten hypotheses with decision criteria (Section~\ref{sec:protocol}, Table~\ref{tab:hypdef}). For RQ1, we expected the memorization term to exceed its permutation null on the training set but not on held-out samples (H2) and to agree with the memorized fraction (H3). For RQ2, we expected the peaks of $V$ and of a label-free heterogeneity term to fall within the memorization phase (H4); stopping at the peak of $\Cb$ to lose less test accuracy than simple baselines (H5); per-sample complexity to be non-inferior to established scores for detecting corrupted samples (H6); the ability of $\Cb$ to distinguish noisy from clean runs to vanish after interpolation (H7); and the peak of $\Cb$ and the trajectories in the entropy-disequilibrium plane to be robust to the choice of functional, to temperature and to the seed (H9, H10). H8 concerns the differences between noise models at complete corruption, and H1 is an implementation check. None of the hypotheses specified what $\Cb$ tracks. The analysis of this question (Section~\ref{sec:mechanism}) is exploratory and was motivated by the outcome of H4 and by the comparison of the H5 rule with established stopping rules.

\paragraph{Contributions.}
This study is primarily empirical. Its theoretical component consists of standard identities and a Fano-type bound formulated for the setting considered here. We compute the signal, memorization and heterogeneity terms exactly for the empirical distribution defined by the logged predictions and the clean and noisy labels (Section~\ref{sec:theory}). At the population level, the memorization term vanishes when the noisy label is conditionally independent of the input and the trained parameters given the clean class, and we give its value under complete memorization. To relate memorization information to the memorized fraction $M$, the fraction of corrupted samples predicted with their noisy label, we derive a Fano-type lower bound on the normalized memorization term $\MI$ in terms of training accuracies on clean and corrupted samples (Proposition~\ref{prop:fano}). We also show that $M=0$ is compatible with $\MI=1$. In our runs, the bound was close to the observed $\MI$, suggesting that training accuracies accounted for much of its measured value. We further quantify the chance level of $M$. A chance correction removed most of the resulting offset in estimated memorization times, while $\MI$ has a zero population null under the stated independence condition (Section~\ref{sec:mi_vs_m}).  We then examine what mean complexity tracked during training and assess its use as a practical tool. Under noise drawn uniformly over the incorrect classes (\textsf{UI}, Section~\ref{sec:setting}), the peak of $\Cb$ preceded the onset of memorization for corruption rates $\rho\leq 0.4$. For $\rho\geq 0.6$, it occurred within the memorization phase, close to the epoch at which mean training confidence reached approximately $0.6$. In these runs, the peak was not temporally distinct from that phase (Section~\ref{sec:mechanism}). As a stopping rule, $\Cb$ performed comparably to the established rules considered for $\rho\leq 0.4$, but incurred substantially larger losses for $\rho\geq 0.6$ and was also outperformed under pair-flip noise. For corrupted-sample detection, time-averaged per-sample complexity performed less well than the established scores evaluated, including simple label-free scores computed from the same predictions (Section~\ref{sec:tools}).

Finally, we examine differences among the noise models. Under complete \textsf{UI} corruption ($\rho=1$), every observed training label differs from the true class and is therefore a complementary label~\citep{Ishida2017}. In our experiments, test predictions retained information about the true class despite below-chance accuracy. This contrasted with the results obtained under the two mechanisms that generate labels independently of the inputs, uniform resampling over all classes (\textsf{SU}) and permutation of the selected labels (\textsf{PERM}) (Section~\ref{sec:noise}).

The study comprises a main investigation on MNIST and an exploratory extension to Fashion-MNIST~\citep{Xiao2017}, pair-flip noise~\citep{Han2018} and four training variants (SGD, a cosine learning-rate schedule, weight decay and training for 400 epochs) (Sections~\ref{sec:protocol} and~\ref{sec:robust}), all with the same architecture. We report both supported and unsupported outcomes. Section~\ref{sec:setting} defines the noise models and the notation used throughout (Tables~\ref{tab:noise} and~\ref{tab:notation}). Source code is available at: \url{https://github.com/NBC777/memorization-info-budget}

\section{Setting and definitions}
\label{sec:setting}
\subsection{Data and label noise}
\label{sec:data}

Let $\mathcal{T}=\{(x_i,y_i,\tilde y_i)\}_{i=1}^n$ be the training set, with clean labels $y_i$ and training (noisy) labels $\tilde y_i$, both in $\{1,\dots,N\}$; $N=10$ for both datasets used here, and $n_y$ denotes the number of training samples of clean class $y$. The \emph{corruption rate} $\rho\in [0,1]$ is the fraction of samples of each clean class selected for relabelling: $\operatorname{round}(\rho\, n_y)$ samples of class $y$ are selected by stratified sampling and relabelled with one of the four mechanisms in Table~\ref{tab:noise}. \textsf{UI} noise at $\rho=1$ turns every label into a complementary label, one that differs from the true class~\citep{Ishida2017}. \textsf{SU} noise produces the uniformly random labels of \citet{Zhang2021}. Because the new label can coincide with the clean one, its effective corruption rate is $\rho(N-1)/N$ in expectation. Under \textsf{PERM}, a permuted label can also coincide with the clean one, so $r\le\rho$. \textsf{PF} is the asymmetric noise used by \citet{Han2018}. We use it only with $\rho<0.5$, so that the clean label remains the majority label within each clean class.

\begin{table}[t]
\caption{\textit{Noise models}. For each selected sample of clean class $y$, the training label $\tilde y$ is generated as shown; unselected samples keep $\tilde y=y$. The corrupted fraction $r$ is the fraction of training samples with $\tilde y\neq y$.}
\label{tab:noise}
\begin{center}
\small
\setlength{\tabcolsep}{4pt}
\begin{tabular}{p{0.23\linewidth} p{0.38\linewidth} c l}
\multicolumn{1}{c}{\bf NAME}  &\multicolumn{1}{c}{\bf RULE FOR A SELECTED SAMPLE}  &\multicolumn{1}{c}{\bf $r$}  &\multicolumn{1}{c}{\bf RATES USED}
\\ \hline \\
\textsf{UI} (uniform over incorrect classes) & $\tilde y$ drawn uniformly from $\{1,\dots,N\}\setminus\{y\}$ & $\rho$ & main grid, extension\\
\textsf{SU} (symmetric uniform) & $\tilde y$ drawn uniformly from all $N$ classes, so it may equal $y$ & $\approx\rho(N-1)/N$ & main grid, $\rho=1$\\
\textsf{PERM} (permutation) & the labels of all selected samples are randomly permuted among themselves & $\le\rho$ & main grid, $\rho=1$\\
\textsf{PF} (pair-flip) & $\tilde y=(y+1)\bmod N$ & $\rho$ & extension, $\rho\le0.4$\\
\hline
\end{tabular}
\end{center}
\end{table}

We write $c_i=\mathds{1}[\tilde y_i\ne y_i]$, $\mathcal{N}=\{i:c_i=1\}$ for the corrupted samples, $\mathcal{L}= \{ i:c_i=0 \}$ for the clean samples, and $r=|\mathcal N|/n$ for the corrupted fraction. In the population, the four mechanisms are class-conditional~\citep{Natarajan2013}:

\begin{assumption}[Class-conditional noise]\label{ass:A1}
For a sample $(X,Y,\Yt)$ drawn from the data distribution and the noise mechanism, $\Yt\perp X\mid Y$. For \textsf{PERM}, this holds up to finite-sample dependence.
\end{assumption}

\noindent Below, empirical quantities use $X$ as a training index drawn uniformly from $\mathcal{T}$, so expectations are averages over the training set. Under this empirical measure, $\tilde y_i$ is a function of the index. Assumption~\ref{ass:A1} therefore concerns the population rather than the empirical measure; Section~\ref{sec:mterm} defines the quantity we estimate and its held-out reference.

\subsection{Training dynamics and memorization}
\label{sec:dynamics}

Training runs for a fixed number of epochs; epoch $t=0$ denotes the initialization, and quantities at epoch $t\ge1$ are computed after the $t$-th epoch. Let $z_t(x)\in\mathbb R^N$ be the logits and $p_t(\cdot\mid x)=\operatorname{softmax}(z_t(x))$ the predictive distribution at epoch $t$. Where stated, we also use the temperature-scaled distribution $\operatorname{softmax}(z_t(x)/T)$, computed after training from the stored logits; $T=1$ unless stated otherwise. The \emph{mean training confidence} is $\frac1n\sum_i\max_kp_t(k\mid x_i)$. For a quantity $f(t)$, its \emph{peak time} $t^*(f)$ is the first epoch $t\ge1$ at which $f$ attains its maximum, and $t_{\rm best}$ is the first epoch $t\ge1$ of highest test accuracy against the clean labels.

\begin{definition}[Memorized fraction, chance level and phase]\label{def:M}
Let $\rho>0$. With predictions $\hat y_i(t)=\arg\max_k p_t(k\mid x_i)$ at epoch $t$, define
\[
M(t)=\frac{1}{|\mathcal N|}\sum_{i\in\mathcal N}\mathds{1}[\hat y_i(t)=\tilde y_i],\qquad
R(t)=\frac{1}{|\mathcal N|}\sum_{i\in\mathcal N}\mathds{1}[\hat y_i(t)=y_i].
\]
These are the fractions of corrupted samples predicted with the noisy label (memorized) and with the clean label (retained), respectively, following \citet{LiuELR2020}. Let $m_y(k,t)$ denote the fraction of training samples of clean class $y$ predicted as $k$ at epoch $t$. The chance level of $M$, its value if predictions did not depend on which samples were corrupted, is
\[
M_{\rm chance}(t)=\frac{1}{|\mathcal N|}\sum_{i\in\mathcal N}m_{y_i}(\tilde y_i,t).
\]
The chance-corrected version, in the spirit of Cohen's $\kappa$~\citep{Cohen1960}, is $\kappa(t)=(M-M_{\rm chance})/(1-M_{\rm chance})$. As pre-specified, the onset $t_{\mathrm{on}}$ is the first epoch at which $M(t)$ attains its minimum, $t_q=\min\{t\ge t_{ \mathrm{on}}:M(t)\ge q\}$, written $t_{10}$, $t_{50}$ and $t_{90}$ for $q=0.1$, $0.5$ and $0.9$, and the memorization phase is $[t_{10},t_{90}]$. Analogous threshold times based on $\kappa$ and on $\MI$ (Section~\ref{sec:mterm}) are reported in Section~\ref{sec:mi_vs_m}. A run interpolates at the first epoch at which its accuracy against the noisy training labels reaches 99.9\%.
\end{definition}

\subsection{Per-sample functionals}
\label{sec:functionals}
For a probability vector $p\in\Delta^{N-1}$, let $u$ denote the uniform distribution and define
\[
S(p)=-\frac{1}{\ln N}\sum_k p_k\ln p_k,\qquad
\mathcal{D}_2(p)=\frac{\|p-u\|_2^2}{(N-1)/N},\qquad
\mathcal{C}(p)=S(p)\,\mathcal{D}_2(p).
\]
All three quantities lie in $[0,1]$; for $N=10$, $\mathcal C\le0.2519$ (Remark~\ref{rem:geom}). We also computed $\mathcal{D}_1(p)=\|p-u\|_1/(2(1-1/N))$ and the Jensen-Shannon disequilibrium $\mathcal D_{JS}$, normalized by its value at a vertex (Appendix~\ref{app:geometry}).

For a run at epoch $t$, write $p_i=p_t(\cdot\mid x_i)$ and $\pbar=\frac1n\sum_ip_i$. Bars denote averages of per-sample quantities: $\Sb=\frac1n\sum_iS(p_i)$, $\Db_2=\frac1n\sum_i\mathcal{D}_2(p_i)$ and $\Cb = \frac 1n \sum_i\mathcal{C}(p_i)$, the \emph{mean complexity}. These averages should not be confused with functionals of the mean prediction, such as $S(\pbar)$ and $\mathcal{D}_2 (\pbar)$. In particular, $\Cb\neq\Sb\,\Db_2$ in general. Finally, $V = \frac 1N \sum_{k=1}^N \operatorname{Var}\{\mathcal{C}(p_i):\tilde y_i=k\}$ is the mean, over noisy-label groups, of the within-group variance of $\mathcal{C}$. We considered $V$ because, within a noisy-label group, clean and corrupted samples may be fitted at different times, which would appear as dispersion of $\mathcal C$ within the group (Observation~\ref{obs:mixture}).

\subsection{The memorization term and its normalizations}
\label{sec:mterm}
Let $\Yh\sim p_t(\cdot\mid X)$ denote soft predictions. By the data-processing inequality, $\Yh$ carries at most the information carried by the vector $p_t(\cdot\mid X)$; we use $\Yh$ because the resulting quantities are exact finite sums and close the budget below. The hard version uses $\hat y(t)$ instead and is invariant to any strictly increasing transformation of the logits, including temperature scaling. We call $I(\Yh;\Yt\mid Y)$ the memorization term, $\iota_{\rm mem}=I(\Yh;\Yt\mid Y)/\ln N$ its normalization in the budget below, and
\[
\MI=\frac{I(\Yh;\Yt\mid Y)}{H(\Yt\mid Y)}\in[0,1]
\]
the memorization information. Two versions are relevant. The \emph{in-sample} value $\MI^{\rm in}$ is computed on the training set itself, with $\theta=\theta(\mathcal T)$, $X$ a training index and $\tilde y_i$ the training labels; it is an exact finite sum over the logged predictions, with $H(\Yt\mid Y)$ from the empirical label channel. It is the quantity we use to measure memorization, that is, the dependence of $\theta(\mathcal T)$ on the realized training labels. Under the empirical measure, $\MI^{\rm in}$ cannot by itself separate memorization of individual labels from learning structure in the noise. Under Assumption~\ref{ass:A1}, however, the noisy label carries no information about the input beyond the clean class, so in the population there is no such structure to learn from the inputs, and a positive in-sample value reflects the dependence of $\theta$ on the realized labels. The \emph{held-out} value $\MI^{\rm out}$ is computed in the same way on test samples whose noisy labels were drawn after training with the same mechanism. Its population value is zero under Assumption~\ref{ass:A1} for any $\theta$ (Proposition~\ref{prop:null}), so the observed $\MI^{\rm out}$ reflects the finite-sample bias of the plug-in estimate. These held-out samples were not used in training and their noisy labels come from the same mechanism; they differ from held-out training samples in that MNIST test images were written by a set of writers disjoint from that of the training images~\citep{LeCun1998}. We report soft and hard versions throughout: the soft version closes the budget of Proposition~\ref{prop:budget} and is the one used in the pre-specified hypothesis H3, whereas the hard version is invariant to temperature and is the more conservative reference for memorization times (Section~\ref{sec:mi_vs_m}). When not stated otherwise, $\MI$ denotes the soft in-sample value $\MI^{\rm in}$. Note that $\MI$ is an information quantity, not a fraction of samples: $\MI=0.5$ does not mean that half of the corrupted samples are memorized.

\begin{table}[t]
\caption{\textit{Notation}. Symbols used throughout the paper and where they are defined.}
\label{tab:notation}
\begin{center}
\small
\begin{tabular}{l p{0.62\linewidth} l}
\multicolumn{1}{c}{\bf SYMBOL}  &\multicolumn{1}{c}{\bf MEANING}  &\multicolumn{1}{c}{\bf DEFINED IN}
\\ \hline \\
$N$, $n$, $n_y$ & number of classes; training-set size; training samples of clean class $y$ & Section~\ref{sec:data}\\
$y$, $\tilde y$, $\hat y$ & clean, noisy (training) and predicted label; $Y$, $\Yt$, $\Yh$ as random variables & Sections~\ref{sec:data}, \ref{sec:mterm}\\
$\rho$, $r$ & corruption rate; fraction of training samples with $\tilde y\ne y$ & Section~\ref{sec:data}\\
\textsf{UI}, \textsf{SU}, \textsf{PERM}, \textsf{PF} & noise models & Table~\ref{tab:noise}\\
$\mathcal N$, $\mathcal L$ & corrupted and clean training samples & Section~\ref{sec:data}\\
$p_t(\cdot\mid x)$, $p_i$, $\pbar$, $T$ & predictive distribution at epoch $t$; for sample $i$; mean prediction; temperature & Sections~\ref{sec:dynamics}, \ref{sec:functionals}\\
$t^*(f)$, $t_{\rm best}$ & peak time of $f$; epoch of best clean test accuracy & Section~\ref{sec:dynamics}\\
$M$, $R$, $M_{\rm chance}$, $\kappa$ & memorized and retained fractions; chance level of $M$; chance-corrected $M$ & Definition~\ref{def:M}\\
$t_{\rm on}$, $t_q$, $[t_{10},t_{90}]$ & onset, threshold times and memorization phase & Definition~\ref{def:M}\\
$S$, $\mathcal D_2$, $\mathcal D_1$, $\mathcal D_{JS}$, $\mathcal C$ & normalized entropy, disequilibria and statistical complexity of one distribution & Section~\ref{sec:functionals}\\
$\Sb$, $\Db_2$, $\Cb$, $V$ & mean entropy, mean disequilibrium, mean complexity; within-group variance of $\mathcal C$ & Section~\ref{sec:functionals}\\
$\Cb_{\mathcal L}$, $\Cb_{\mathcal N}$ & mean complexity of clean and of corrupted samples & Observation~\ref{obs:mixture}\\
$\MI^{\rm in}$, $\MI^{\rm out}$ & memorization information on training and held-out samples (soft or hard) & Section~\ref{sec:mterm}\\
$\iota_{\rm sig}$, $\iota_{\rm mem}$, $\iota_{\rm het}$ & normalized signal, memorization and heterogeneity terms & Proposition~\ref{prop:budget}\\
$I(\Yh;X\mid\Yt)$ & heterogeneity term without clean labels (label-free) & Proposition~\ref{prop:budget}\\
$K$, $\varphi_K$ & largest support of $\Yt$ given $Y$; Fano function & Proposition~\ref{prop:fano}\\
H1-H10 & pre-specified hypotheses & Table~\ref{tab:hypdef}\\
\hline
\end{tabular}
\end{center}
\end{table}

\section{Theory}
\label{sec:theory}

The first two results are standard identities, stated in the form we use here.

\begin{proposition}[Aggregation gaps]\label{prop:jensen}
\[
S(\pbar)-\Sb=\frac{I(X;\Yh)}{\ln N}\;\ge 0,\qquad
\Db_2-\mathcal{D}_2(\pbar)=\frac{\frac1n\sum_i\|p_i-\pbar\|_2^2}{(N-1)/N}\;\ge 0 .
\]
The first gap is the mutual information between inputs and predictions; an analogous entropy difference, with averaging over model uncertainty rather than inputs, is used in Bayesian active learning~\citep{Houlsby2011}. Both gaps are Bregman informations of the cloud $\{p_i\}$, associated with negative Shannon entropy and the squared Euclidean norm, respectively~\citep{BanerjeeBregman2005}. In the $(S,\mathcal{D}_2)$ plane, the vector from $(S(\pbar),\mathcal{D}_2(\pbar))$ to the centroid $(\Sb,\Db_2)$ is therefore 
\[
\big(-I(X;\Yh)/\ln N,\ +\tfrac{1}{n}\sum_i\|p_i-\pbar\|^2/((N-1)/N)\big).
\]
\end{proposition}
\begin{proof}
Since $H(\Yh)=H(\pbar)$ and $H(\Yh\mid X)=\frac1n\sum_iH(p_i)$, their difference is $I(X;\Yh)$. Dividing by $\ln N$ gives the first identity. For the second, $\|p-u\|^2=\|p\|^2-1/N$, hence 
\[
 \Db_2-\mathcal D_2(\pbar) = \frac{ \frac1n\sum_i\|p_i\|_2^2-\|\pbar\|_2^2 }{(N-1)/N} = \frac{\frac1n\sum_i\|p_i-\pbar\|_2^2} {(N-1)/N}. 
\]
\end{proof}

\begin{proposition}[Information budget]\label{prop:budget} Fix a collection of predictive distributions
$p_i\in\Delta^{N-1}$, with $N\geq2$, and their associated
clean and noisy labels $(y_i,\tilde y_i)$.
Under the empirical distribution, let $X$ be uniform over
the sample indices $\{1,\dots,n\}$, set $Y=y_X$ and
$\Yt=\tilde y_X$, and define
$\Yh\mid X=i\sim\mathrm{Categorical}(p_i)$.
\begin{equation}
\underbrace{S(\pbar)}_{\text{marginal}}=\underbrace{\Sb}_{\text{uncertainty}}+\underbrace{\frac{I(\Yh;Y)}{\ln N}}_{\iota_{\rm sig}}+\underbrace{\frac{I(\Yh;\Yt\mid Y)}{\ln N}}_{\iota_{\rm mem}}+\underbrace{\frac{I(\Yh;X\mid Y,\Yt)}{\ln N}}_{\iota_{\rm het}} .
\label{eq:budget}
\end{equation}
Here $\iota_{\rm het}$ measures the dispersion of predictions within each $(y,\tilde y)$ cell. Without clean labels, $S(\pbar)=\Sb+I(\Yh;\Yt)/\ln N+I(\Yh;X\mid\Yt)/\ln N$. The quadratic analogue is
\[
\tfrac1n\textstyle\sum_i\|p_i-\pbar\|^2=\sum_y\pi_y\|\pbar_y-\pbar\|^2+\sum_{y,\tilde y}\pi_{y\tilde y}\|\pbar_{y\tilde y}-\pbar_y\|^2+\sum_{y,\tilde y}\pi_{y\tilde y}\,\overline{\|p_i-\pbar_{y\tilde y}\|^2}^{\,(y,\tilde y)},
\]
where $\pi_y$ and $\pi_{y\tilde y}$ are the empirical frequencies, $\pbar_y$ and $\pbar_{y\tilde y}$ are group means, and the final bar denotes the mean within the cell.
\end{proposition}
\begin{proof}
We have $H(\Yh)=H(\Yh\mid X)+I(X;\Yh)$. Under the empirical measure, $(Y,\Yt)$ are functions of the index $X$, so $I(X;\Yh)=I(X,Y,\Yt;\Yh)$. The chain rule then gives $I(Y,\Yt,X;\Yh)=I(Y;\Yh)+I(\Yt;\Yh\mid Y)+I(X;\Yh\mid Y,\Yt)$~\citep{CoverThomas2006}. The quadratic identity follows from the law of total variance applied to the nested groupings. 
\end{proof}

The identities hold exactly for any predictive distribution. For the soft version, the values of the terms depend on logit scale, which is one reason to report the hard version alongside it.

\begin{proposition}[Null of the memorization term]\label{prop:null}
Assume that, conditional on $Y$, the noisy label is independent of the input and the trained parameters jointly: $\Yt\perp(X,\theta)\mid Y$. Under Assumption~\ref{ass:A1}, this holds for a sample whose noisy label was drawn independently of training, such as a held-out sample, and, up to finite-sample dependence, for a training label permuted within its clean class, because the permutation is drawn independently of the training run that produced $\theta$. Then the population value of $I(\Yh;\Yt\mid Y)$ is zero.
\end{proposition}
\begin{proof}
Conditional on $Y$, $\Yh$ is generated from $(X,\theta)$ and $\Yt\perp(X,\theta)\mid Y$. Hence, $\Yt\perp\Yh\mid Y$.
\end{proof}

On a finite sample, the plug-in estimate is biased upwards~\citep{Paninski2003}, and the bias depends on the cell sizes. We therefore compare it with a permutation null obtained by permuting $\tilde y$ within each clean class. This tests the exchangeability of noisy labels within classes conditional on the predictions. Under Assumption~\ref{ass:A1}, a value above this null on the training set indicates dependence of $\theta$ on the realized noise, that is, memorization. Under instance-dependent noise, it may also reflect learnable patterns in the noise, and a liberal test can produce false rejections (Section~\ref{sec:valid}).

\begin{proposition}[End point under complete memorization]\label{prop:endpoint}
If $p_i=\delta_{\tilde y_i}$ for every $i$, then $\Sb=0$, $\iota_{\rm het}=0$, and
\[
\iota_{\rm sig}=\frac{I(\Yt;Y)}{\ln N},\qquad \iota_{\rm mem}=\frac{H(\Yt\mid Y)}{\ln N},\qquad \MI=1 .
\]
The same holds for the hard budget whenever $\hat y_i=\tilde y_i$ for all $i$.
\end{proposition}
\begin{proof}
Since $\Yh=\Yt$ almost surely, $H(\Yh\mid X)=0$, $I(\Yh;Y)=I(\Yt;Y)$, $I(\Yh;\Yt\mid Y)=H(\Yt\mid Y)$, and $I(\Yh;X\mid Y,\Yt)=0$.
\end{proof}

\noindent Proposition~\ref{prop:endpoint} follows from the definitions, so we use it as a sanity check rather than as evidence. In what follows, reference values are computed from the empirical label channel. For \textsf{UI} noise with exactly balanced classes, $H(\Yt\mid Y)=h(\rho)+\rho\ln(N-1)$, where $h$ is the binary entropy.  For $N=10$, the corresponding values of $(\iota_{\rm sig},\iota_{\rm mem})$ are approximately $(0.592,0.408)$, $(0.326, 0.674)$, $(0.135, 0.865)$, $(0.019, 0.981)$ and $(0.046, 0.954)$ for $\rho=0.2, 0.4, 0.6, 0.8, 1$, respectively. MNIST classes are not exactly balanced, with  5\,421 to 6\,742 training images per class, and the empirical values differ from these by at most one unit in the third decimal; for example, they are $0.020$ and $0.980$ at $\rho=0.8$. Away from the end point, $\MI$ can be related to the training accuracies. Let $\mathcal S_y$ denote the support of $\Yt$  conditional on  $Y=y$, let $K=\max_y|\mathcal S_y|$, and  define $\varphi_K(e)=h(e)+e\ln(K-1)$. This function is concave on $[0,1]$ and nondecreasing on $[0,(K-1)/K]$ and it attains its maximum $\ln K$ at $e=(K-1)/K$.

\begin{proposition}[Fano lower bound]\label{prop:fano}
For any prediction $\Yh$, whether hard ($\Yh=\hat y(X)$) or soft ($\Yh\sim p(\cdot\mid X)$), let $e=P(\Yh\neq\Yt)$ denote its error with respect to the training labels. Then
\[
\MI\;\ge\;1-\frac{\varphi_K\!\big(\min\{e,(K-1)/K\}\big)}{H(\Yt\mid Y)},\qquad e=1-(1-r)A_{\mathcal L}-r\,M,
\]    
where, for hard predictions, $A_{\mathcal L}$ is the accuracy on clean samples and $M$ is the memorized fraction. For soft predictions, $A_{\mathcal L}$ and $M$ are the mean probabilities assigned to the clean label on clean samples and to the noisy label on corrupted samples, respectively. There is no upper bound on $\MI$ in terms of $M$: for $N\ge3$, a prediction that is an injective function of $\Yt$ conditional on $Y$, but never equals $\Yt$ on corrupted samples, has $M=0$ and $\MI=1$.
\end{proposition}
\begin{proof}
Fix $y$ and define an estimator $g(y,\hat y)$ of $\Yt$ taking values in $\mathcal S_y$: $g=\hat y$ if $\hat y\in\mathcal S_y$; otherwise, set $g=y$ if $y\in\mathcal S_y$; otherwise, set $g=s_y$, where $s_y$ is a fixed element of $\mathcal S_y$. The last case occurs only when every  sample of class $y$ is corrupted, as under \textsf{UI} noise at $\rho=1$. By the data-processing inequality and Fano's inequality~\citep[Thm.~2.10.1]{CoverThomas2006}, $H(\Yt\mid \Yh,Y=y)\le H(\Yt\mid g(y,\Yh),Y=y)\le\varphi_{|\mathcal S_y|}(e_y)\le\varphi_K(e_y)$, where $e_y=P(g(Y,\Yh)\ne\Yt\mid Y=y)$; classes with $|\mathcal S_y|=1$ contribute zero. Averaging over $y$ with weights $\pi_y$ and using concavity gives $H(\Yt\mid Y,\Yh)\le\varphi_K(\bar e)$ with $\bar e=\sum_y\pi_ye_y$. 

On a corrupted sample, $g=\Yt$ whenever $\Yh=\Yt$, because then $\Yh\in\mathcal S_y$. On a clean sample, $g=\Yt=y$ whenever $\Yh=y$. Replacing a prediction outside $\mathcal S_y$ cannot turn  a correct prediction into an error. Hence $\bar e\le 1-(1-r)A_{\mathcal L}-rM=e$. Since $\varphi_K$ is nondecreasing up to $(K-1)/K$ and attains its maximum there, $\varphi_K(\bar e)\le\varphi_K(\min\{e,(K-1)/K\})$. Finally, $I(\Yh;\Yt\mid Y)=H(\Yt\mid Y)-H(\Yt\mid Y,\Yh)$; dividing by $H(\Yt\mid Y)$ gives the bound. All steps hold for the empirical joint distribution, so the bound  also applies to the plug-in quantities. For the counterexample, let $\Yh=\tau_Y(\Yt)$, where each $\tau_y$ is injective, $\tau_y(y)=y$ and $\tau_y(\tilde y)\ne\tilde y$ for $\tilde y\ne y$ (for instance a cyclic shift of the $N-1$ labels different from $y$, which needs $N\ge3$). Then $H(\Yt\mid Y,\Yh)=0$, so $\MI=1$, while $\Yh\ne\Yt$ on every corrupted sample, so $M=0$.
\end{proof}

The bound shows that agreement with the noisy labels forces memorization information: $\MI\to1$ as $M\to1$ and $A_{\mathcal L}\to1$. Conversely, $\MI$ can also register dependence on the realized noise that does not take the form of agreement. The difference between $\MI$ and its lower bound has two sources: slack in Fano's inequality, and corrupted samples for which neither the clean nor the noisy label is predicted. For pair-flip noise ($K=2$), the bound reduces to $\MI\ge1-h(e)/h(\rho)$ for $e\le1/2$. At low pair-flip rates the normalization $H(\Yt\mid Y)=h(\rho)$ is small ($0.325$ nats at $\rho=0.1$), so small absolute changes in $I(\Yh;\Yt\mid Y)$ produce larger changes in $\MI$.

We next record an elementary identity used to interpret the complexity summaries.

\begin{observation}[Decomposition of the mean complexity]\label{obs:mixture}
Let $\Cb_{\mathcal L}$ and $\Cb_{\mathcal N}$ denote the mean complexities of clean and corrupted samples, respectively. Then $\Cb=(1-r)\,\Cb_{\mathcal L}+r\,\Cb_{\mathcal N}$. Within noisy-label group $k$, let $r_k$ be the corrupted fraction, let $\Cb_{\mathcal L,k}$ and $\Cb_{\mathcal N,k}$ be the mean complexities of its clean and corrupted samples, let $V_{\mathcal L,k}$ and $V_{\mathcal N,k}$ be the corresponding variances, and let $\operatorname{Var}_k$ be the variance of $\mathcal C$ over the whole group, so that $V=\frac1N\sum_k\operatorname{Var}_k$. The law of total variance gives $\operatorname{Var}_k=r_kV_{\mathcal N,k}+(1-r_k)V_{\mathcal L,k}+r_k(1-r_k)(\Cb_{\mathcal N,k}-\Cb_{\mathcal L,k})^2$. By the Bhatia-Davis inequality~\citep{BhatiaDavis2000} and the concavity of $x(\mathcal{C}_{\max}-x)$, $V\le\Cb_{\rm mac}(\mathcal{C}_{\max}-\Cb_{\rm mac})$, where $\Cb_{\rm mac}$ is the mean, over groups, of the group means of $\mathcal{C}$. This bound is used only to normalize $V$ in an appendix figure.
\end{observation}

\noindent The first identity holds for the average of any per-sample statistic, so it does not by itself explain the behavior of $\Cb$. We use it only to identify which samples contribute to $\Cb$ at a given epoch.

\begin{remark}[Geometry of the plane]\label{rem:geom}
The image of the simplex under $(S,\mathcal{D}_2)$ is a compact region. Since $\mathcal{D}_2$ is affine in $\sum_kp_k^2$, its boundaries correspond to those of the entropy-index-of-coincidence diagram~\citep{HarremoesTopsoe2001}. We evaluated these boundaries numerically using parametric families containing the extremal distributions (Appendix~\ref{app:geometry}). For $N=10$, the numerical maximum of $\mathcal{C}$ is $0.2519$, attained at a two-level distribution whose largest probability is approximately 0.76; the limit as $N\to\infty$ is  $4/27\approx 0.148$. The centroid $(\Sb,\Db_2)$ lies in the convex hull of the region but need not lie in the region itself.
\end{remark}

\section{Experimental protocol}
\label{sec:protocol}

\paragraph{Model and training.} We used MNIST~\citep{LeCun1998}, comprising 60\,000 training and 10\,000 test images, with inputs standardized with a mean of 0.1307 and  a standard deviation of 0.3081. The model was a 784 - 512 - 512 - 10 multilayer perceptron with ReLU activations, no dropout or weight decay, and 669\,706 parameters. It was trained for 200 epochs using Adam~\citep{Kingma2015}, a constant learning rate of $10^{-3}$, a batch size of 256, and cross-entropy loss on the noisy labels. Deterministic GPU settings were enabled. The label-noise seed was separate from the seed used for initialization and mini-batch ordering.

\paragraph{Grid (89 runs).} The grid comprised \textsf{UI} noise with $\rho\in\{0,0.2,0.4,0.6,0.8,1\}$ and 10 seeds per rate (60 runs); \textsf{SU} and \textsf{PERM} at $\rho=1$ with 10 seeds each (20 runs); and a $3\times 3$ design under \textsf{UI} noise at $\rho=0.4$, crossing three noise realizations with three initialization seeds (9 runs). The main grid (80 runs) used noise seed $=10\,000+$ initialization seed. The nine crossed runs were analyzed separately. Training used PyTorch 2.6.0 (CUDA 12.4 build) on a single NVIDIA GeForce RTX 3060 GPU, with a mean wall-clock time of about 1.6-1.9 minutes per run.

\paragraph{Exploratory extension (280 runs).} The extension used the same architecture and logging protocol, with 10 seeds per condition and noise seed $=10\,000+$ seed; together with the main grid, the study comprises 369 runs. It included pair-flip noise on MNIST with $\rho \in \{0.1, 0.2, 0.3, 0.4\}$ (40 runs) and Fashion-MNIST~\citep{Xiao2017} with \textsf{UI} noise at the six rates of the main grid and PF noise at $\rho \in \{0.2, 0.4\}$ (80 runs). Fashion-MNIST inputs were standardized using the mean and standard deviation of the training pixels.

We evaluated four training variants on MNIST under \textsf{UI} noise. SGD with momentum $0.9$ and a learning rate of $0.01$ was evaluated at all six noise rates, comprising 60 runs. Adam with a cosine learning-rate schedule and Adam with weight decay $10^{-4}$ were each evaluated at $\rho\in\{0.2,\dots,0.8\}$, comprising 40 runs per variant. The fourth variant extended Adam training to 400 epochs at $\rho\in\{0.8,1\}$, comprising 20 runs. Each run was compared with its counterpart under the main protocol, matched by noise rate and seed. These analyses were defined after the main results were obtained and are therefore exploratory.

\paragraph{Logging.} At epoch 0 (initialization) and after every epoch, the model was evaluated in evaluation mode on the full training and test sets. At every epoch, we stored scalar metrics computed from float32 logits, per-sample argmax predictions, margins and losses. Full logits were stored every epoch through epoch 30, every two epochs through epoch 100, and every five epochs thereafter. The permutation null of $I(\Yh;\Yt\mid Y)$, based on 100 within-class permutations, was computed at 37 scheduled epochs. For the test set, fresh noisy labels were drawn with the same mechanism and rate and used only for the held-out null check and the noisy-validation baseline.

\paragraph{Pre-specified hypotheses.}
Ten hypotheses and their decision criteria (Table~\ref{tab:hypdef}) were fixed in a written protocol before the main grid was run. We describe them as pre-specified rather than pre-registered because the protocol was not deposited in a public registry. The written protocol accompanied the code before the first run of the main grid was completed and was placed under version control the following day, without changes. An earlier pilot implementation, which used a different architecture and logging protocol, informed the hypotheses. None of its results are reported here. Table~\ref{tab:deviations} lists deviations from the plan. H1 is an implementation check rather than a scientific hypothesis. H1 - H3 concern the information budget and the memorization term (RQ1). H4 - H7, H9, and H10 concern the complexity summaries and their use as tools (RQ2). H8 concerns the noise models. The expectations underlying these hypotheses are summarized in Section~\ref{sec:intro}.

\begin{table}[t]
\caption{\textit{Pre-specified hypotheses and decision criteria}. $q_{95}$ and $q_{99}$ are quantiles of the within-class permutation null; $I(\Yh;X\mid\Yt)$ is the heterogeneity term computed without clean labels (Proposition~\ref{prop:budget}).}
\label{tab:hypdef}
\begin{center}
\begin{tabular}{l p{0.45\linewidth} p{0.40\linewidth}}
\multicolumn{1}{c}{\bf }  &\multicolumn{1}{c}{\bf HYPOTHESIS}  &\multicolumn{1}{c}{\bf DECISION CRITERION}
\\ \hline \\
H1 & The identities of Propositions~\ref{prop:jensen} and~\ref{prop:budget} hold numerically & maximum absolute residual $<10^{-8}$\\
H2 & $I(\Yh;\Yt\mid Y)$ is within its null on test data with fresh noisy labels, and above its null on training data after $t_{10}$ & test: within $q_{95}$ in $\ge 90\%$ of scheduled epochs; training: above $q_{99}$ in $\ge 95\%$ of scheduled epochs after $t_{10}$\\
H3 & $M(t)$ and $\MI(t)$ agree (\textsf{UI}, $\rho \in \{0.2,\dots,0.8\}$) & median Spearman correlation over epochs $\ge 0.9$\\
H4 & The peaks of $V$ and of $I(\Yh;X\mid\Yt)$ fall within $[t_{10},t_{90}]$ & $\ge 80\%$ of runs, for both\\
H5 & Stopping at $\arg\max_{t\ge1}\Cb$ loses less test accuracy than the better of the final epoch and the loss plateau & 95\% CI of the paired difference $<0$\\
H6 & The per-sample $\mathrm{mean}_t\,\mathcal{C}_i$ (first 50 epochs) is non-inferior to the best baseline for detecting corrupted samples & lower bound of the 95\% CI of $\Delta$AUROC $> -0.02$\\
H7 & After interpolation, the run-level AUROC of $\Cb$ between $\rho=0$ and $\rho>0$ runs falls & final AUROC $\le 0.6$\\
H8 & At $\rho=1$, \textsf{UI} gives below-chance test accuracy with $\iota_{\rm sig}^{\rm test}>0.01$, while \textsf{SU} and \textsf{PERM} give chance accuracy with $\iota_{\rm sig}^{\rm test}<0.01$ & \textsf{UI} accuracy $<10\%$; \textsf{SU}, \textsf{PERM} within $10\pm3\%$; all conditions\\
H9 & The peak time of $\Cb$ is robust to the functional ($\mathcal D_1$, $\mathcal D_{JS}$) and to temperature (0.5, 2) & Kendall $\tau\ge0.8$ (if $\tau$ is undefined: median $|\Delta t|\le2$ epochs)\\
H10 & Centroid trajectories $(\Sb(t),\Db_2(t))$ are stable across seeds & largest mean within-rate / smallest mean between-rate discrete Fr\'echet distance $<0.5$\\
\hline
\end{tabular}
\end{center}
\end{table}

\paragraph{Statistics.} The statistical unit is the run. Confidence intervals (CI) are percentile bootstrap intervals over runs, based on 10\,000 resamples and computed separately for each condition; with 10 runs per condition they may undercover and are best read as descriptive. Intervals are written as mean [lower, upper]. As pre-specified, H5 was also tested with a one-sided paired Wilcoxon signed-rank test over the 40 runs with $\rho \in \{0.2,\dots,0.8\}$ and Holm's correction was applied to the p-values of H5 and H6, treated as one family of two tests. In the exploratory per-rate comparisons, Holm's correction was applied over the four values of $\rho$ within each comparison. Runs with the same seed at different $\rho$ share the initialization, the mini-batch order and the noise seed, so this pooled test treats dependent runs as independent; per-$\rho$ tests over independent runs are reported in Section~\ref{sec:tools}. The ten seeds were 42, 123, 456, 789, 101112, 2024, 7, 314, 2718 and 1618. Because the noise seed was tied to the initialization seed in the main grid, their effects are confounded there; the $3\times3$ design separates them descriptively. No multiplicity correction was applied to the exploratory analyses, which should be interpreted as descriptive.

Trends in event times with $\rho$ were summarized using Spearman correlations and per-$\rho$ values. Ordinary least squares with standard errors clustered by initialization seed (10 clusters, CR1 correction, $t$ reference distribution with 9 degrees of freedom) is reported only as a coarse summary, because the relation is step-like. A linear mixed model yielded the same slopes but converged at the boundary, with an estimated seed variance of approximately zero. 

\begin{table}[t]
\caption{Deviations from the pre-specified plan.}
\label{tab:deviations}
\begin{center}
\begin{tabular}{l p{0.75\linewidth}}
\multicolumn{1}{c}{\bf ITEM}  &\multicolumn{1}{c}{\bf DEVIATION AND REASON}
\\ \hline \\
Memorization times & The pre-specified $t_q$, based on $M$, is retained, but $M$ has a chance level (Section~\ref{sec:mi_vs_m}). Times based on $\kappa$ and $\MI$ are added as exploratory analyses.\\
H7 & H7 was not testable under the protocol because no corrupted run interpolated. It was examined post hoc on SGD runs (Section~\ref{sec:robust}). The pre-specified criterion was one-sided. We also report the symmetric version.\\
Trend model & The linear mixed model was replaced by OLS with seed-clustered standard errors because of the boundary fit and supplemented with values for each $\rho$.\\
Baselines & Noisy validation and Label Wave were added after the protocol was fixed  and are exploratory.\\
Extension & All 280 extension runs were planned after the main results.\\
\hline
\end{tabular}
\end{center}
\end{table}

\paragraph{Early-stopping evaluation (exploratory).} The test set was split once, using a fixed seed, into a validation half and an evaluation half. Every stopping rule is scored by the loss in clean test accuracy on the evaluation half relative to the best epoch on that half. The rules are: the final epoch; a training-loss plateau, the first epoch at which the noisy-label training loss decreased by less than 1\% over the preceding five epochs; $\arg\max_t\Cb$ (the pre-specified rule) and $\arg\max_tV$; temperature and functional variants of $\Cb$; noisy validation, the accuracy against fresh noisy labels on 5000, 1000 or 500 samples of the validation half; and Label Wave~\citep{YuanLabelWave2024}.

Here a rule is called label-free if it uses neither clean labels nor held-out data. The noisy training labels, which a practitioner has, may be used. Noisy validation uses held-out data with noisy labels generated by the same mechanism, which is information that label-free rules do not have. Label Wave tracks the fraction of training samples whose predicted class changes between epochs, smooths this fraction using a trailing window $k$, and stops at the first minimum with patience $p$. We re-implemented it from its published description using a grid of $k\in\{1,3,5\}$ and $p\in\{3,5,10\}$ with prediction changes computed once per epoch on the full training set. The original algorithm evaluates prediction changes every $n$ training steps and we set this interval to one epoch. Because our settings of $n$, $k$ and $p$ may differ from those of the original implementation, we report the full grid. 

For rule selection, a family of 16 label-free rules was ranked using seeds $\{42,123,456,789,101112\}$ according to their loss on the validation half of the test set. The selected rule was then evaluated on the evaluation half using the other five seeds. The family comprised $\arg\max\Cb$ with $\mathcal D_1$, $\mathcal D_2$, and $\mathcal D_{JS}$ and with temperatures $0.5$ and $2$; $\arg\max V$; the nine Label Wave configurations; and the rule based on the maximum of $H(\hat y\mid\tilde y)$. Selection and evaluation therefore used different seeds and different test images.  The ranking, however, used the clean labels of the validation half. Thus, although the selected rule was label-free, the selection procedure was not. This use of clean labels may favor the selected rule relative to selection based on noisy validation labels. Rule selection using the noisy labels of the validation half, which a practitioner could carry out, was not evaluated.

\section{Results}
\label{sec:results}

\subsection{Overview of the pre-specified outcomes}

\begin{table}[t]
\caption{\textit{Outcomes of the pre-specified hypotheses} (definitions and criteria in Table~\ref{tab:hypdef}). Main grid, 80 runs; H1 and H2 evaluated over all 89 runs, including the nine crossed runs. $\iota_{\rm sig}$ is normalized by $\ln N$.}
\label{tab:hyp}
\begin{center}
\begin{tabular}{l l p{0.38\linewidth} l}
\multicolumn{1}{c}{\bf }  &\multicolumn{1}{c}{\bf SUBJECT}  &\multicolumn{1}{c}{\bf STATISTIC}  &\multicolumn{1}{c}{\bf DECISION}
\\ \hline \\
H1 & identities & maximum absolute residual $=1.2\times10^{-14}$ & pass\\
H2 & null of $I(\Yh;\Yt\mid Y)$ & test: 0.93 of scheduled epochs within $q_{95}$; training: 0.99 above $q_{99}$ (100 permutations) & pass\\
H3 & agreement of $M$ and $\MI$ & median Spearman 0.993 ($n=40$) & pass\\
H4 & peaks of $V$ and $I(\Yh;X\mid\Yt)$ in phase & $V$: 1.00; $I(\Yh;X\mid\Yt)$: 0.55 ($n=40$) & fail\\
H5 & stopping at the peak of $\Cb$ & $\Delta=-16.3$ p.p.\ [$-19.0$, $-13.6$]. Wilcoxon, Holm-adjusted $p=1.8\times10^{-12}$ & pass\\
H6 & per-sample detection & $\Delta$AUROC $=-0.318$ [$-0.381$, $-0.255$] vs.\ EL2N (60 runs; $-0.481$ for \textsf{UI}, $0<\rho<1$) & fail\\
H7 & after interpolation & no corrupted protocol run interpolated. Post hoc (SGD): final AUROC 1.00, criterion not met & n/a\\
H8 & noise models at $\rho=1$ & \textsf{UI} 0.5\%, 0.040; \textsf{SU} 10.1\%, 0.002; \textsf{PERM} 10.1\%, 0.002 & pass\\
H9 & robustness of the peak time & $\mathcal{D}_1$: 0.80; $\mathcal{D}_{JS}$: 0.84 (met); $T{=}0.5$: 0.46; $T{=}2$: 0.75 (not met) & fail\\
H10 & stability across seeds & $R=2.30$ & fail\\
\hline
\end{tabular}
\end{center}
\end{table}

Table~\ref{tab:hyp} summarizes the decisions: five hypotheses were supported (H1, H2, H3, H5 and H8), four were not (H4, H6, H9 and H10) and H7 could not be tested under the protocol. The remainder of this section examines these outcomes alongside exploratory analyses that qualify several of them. Some criteria were lenient: H3 (Section~\ref{sec:valid}), H4 (Section~\ref{sec:mechanism}) and H5, whose baselines, the final epoch and the loss plateau, lost 12-66 percentage points of test accuracy on the evaluation half (Table~\ref{tab:es}). Others were difficult to pass by construction: H6, which compares label-free features with scores that use the noisy label, and H10, which requires seed-to-seed variation to be less than half of the smallest separation between corruption rates. The pass/fail labels should therefore not be weighed equally.  

\subsection{Training dynamics}

\begin{table}[t]
\caption{\textit{Training dynamics under \textsf{UI} noise}.  Means over 10 seeds and standard deviation in parentheses. Test accuracy is measured against the clean labels. Final $M$ is the memorized fraction at epoch 200. Times are in epochs (Definition~\ref{def:M}). $^{*}$At $\rho = 1$, $M$ starts above 0.1 because of its chance level, so $t_{10}$ (0 or 1, depending on the seed) is not a memorization time. The corresponding time is 9.6 when based on $\kappa$ and 15.3 when based on hard $\MI$ (Table~\ref{tab:mi_vs_m}). The best test accuracy and $t_{\rm best}$ exclude epoch 0; at $\rho=1$ test accuracy stayed below chance from epoch 1 onwards, so these two entries are not reported.}
\label{tab:dyn}
\begin{center}
\small
\begin{tabular}{lcccccccc}
\multicolumn{1}{c}{\bf $\rho$}  &\multicolumn{1}{c}{\bf \begin{tabular}[c]{@{}c@{}}BEST TEST\\ACCURACY\end{tabular}}  &\multicolumn{1}{c}{\bf \begin{tabular}[c]{@{}c@{}}FINAL TEST\\ACCURACY\end{tabular}}  &\multicolumn{1}{c}{\bf $t_{\rm best}$}  &\multicolumn{1}{c}{\bf $t_{10}$}  &\multicolumn{1}{c}{\bf $t_{50}$}  &\multicolumn{1}{c}{\bf $t_{90}$}  &\multicolumn{1}{c}{\bf \begin{tabular}[c]{@{}c@{}}FINAL TRAIN ACC.\\\\(NOISY LABELS)\end{tabular}}  &\multicolumn{1}{c}{\bf FINAL $M$}
\\ \hline \\
0   & 0.986 (0.001) & 0.985 (0.001) & 109.1 & -- & -- & -- & 1.000 & --\\
0.2 & 0.973 (0.001) & 0.855 (0.005) & 5.6 & 11.0 & 18.2 & 44.2 & 0.992 & 0.979\\
0.4 & 0.965 (0.002) & 0.659 (0.011) & 4.9 & 12.0 & 20.1 & 44.5 & 0.987 & 0.980\\
0.6 & 0.948 (0.002) & 0.430 (0.009) & 4.1 & 11.8 & 20.9 & 49.8 & 0.979 & 0.974\\
0.8 & 0.863 (0.007) & 0.200 (0.006) & 3.1 & 10.3 & 23.2 & 71.8 & 0.963 & 0.962\\
1.0 & -- & 0.005 (0.001) & -- & 0.5$^{\ast}$ & 20.6 & 64.3 & 0.962 & 0.962\\
\hline
\end{tabular}
\end{center}
\end{table}

\begin{figure}[h]
\begin{center}
\includegraphics[width=\textwidth]{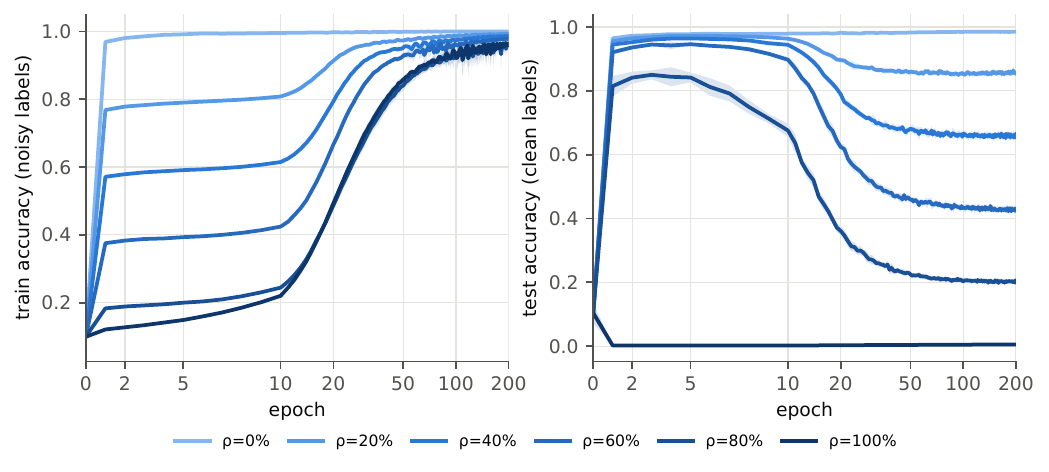}
\end{center}
\caption{Training accuracy against the noisy labels (left) and test accuracy against the clean labels (right) under \textsf{UI} noise. Mean $\pm$ s.d.\ over 10 seeds. The epoch axis is symmetric-logarithmic.}
\label{fig:acc}
\end{figure}

\begin{figure}[h]
\begin{center}
\includegraphics[width=\textwidth]{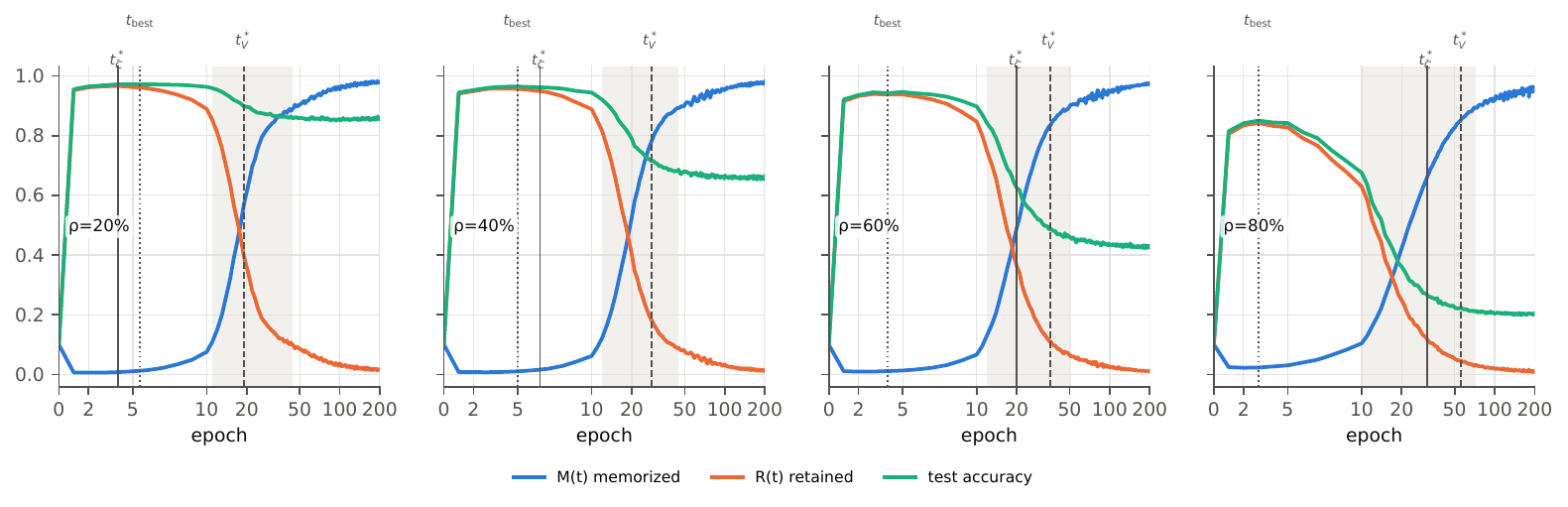}
\end{center}
\caption{Memorized fraction $M(t)$, retained fraction $R(t)$ and test accuracy under \textsf{UI} noise. Shading indicates the memorization phase $[t_{10},t_{90}]$, defined using $M$ (medians). Vertical lines indicate the median of the peaks of $\Cb$ and $V$ and of the best test accuracy.}
\label{fig:phase}
\end{figure}

Figure~\ref{fig:acc} and Table~\ref{tab:dyn} show the two-phase behavior reported in studies of learning with noisy labels~\citep{Arpit2017,LiuELR2020}. At epoch 5, the network predicted the clean label for most corrupted samples: $R$ was $0.963$, $0.957$, $0.938$, and $0.828$ for $\rho=0.2$, $0.4$, $0.6$, and $0.8$, respectively, while $M\leq 0.03$. At the same epoch, training accuracy measured against the noisy labels remained close to $1-\rho$, with respective values of $0.790$, $0.591$, $0.393$, and $0.200$. Memorization, as measured by $M$, began around epochs 10 to 12 and reached its halfway point around epochs 18 to 23  (Figure~\ref{fig:phase}). Test accuracy peaked earlier, between epochs 3 and 6, before the onset of memorization, and subsequently declined as $M$ increased. After 200 epochs, $M$ ranged from $0.96$ to $0.98$, but training had not reached interpolation: none of the runs with corrupted labels attained $99.9\%$ training accuracy under the main protocol, which used Adam with a constant learning rate. This limits the analysis under that protocol; runs that reached interpolation were obtained in the extension (Section~\ref{sec:robust}).

\subsection{The aggregation gap and the budget}

\begin{figure}[h]
\begin{center}
\includegraphics[width=\textwidth]{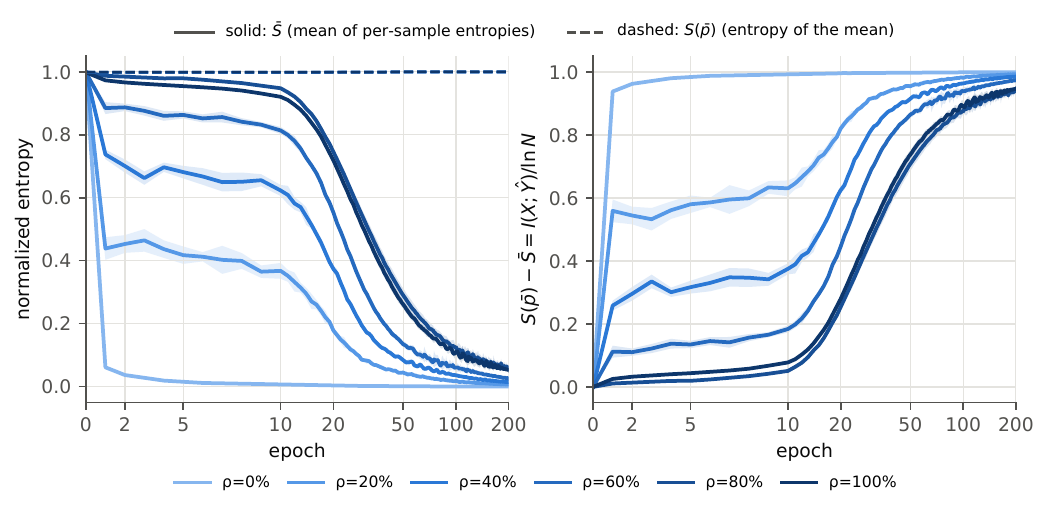}
\end{center}
\caption{Left: mean entropy $\Sb$ (solid) and entropy of the mean prediction $S(\pbar)$ (dashed). Right: the aggregation gap $S(\pbar)-\Sb=I(X;\Yh)/\ln N$ (Proposition~\ref{prop:jensen}). \textsf{UI} noise.}
\label{fig:gap}
\end{figure}

\begin{figure}[h]
\begin{center}
\includegraphics[width=\textwidth]{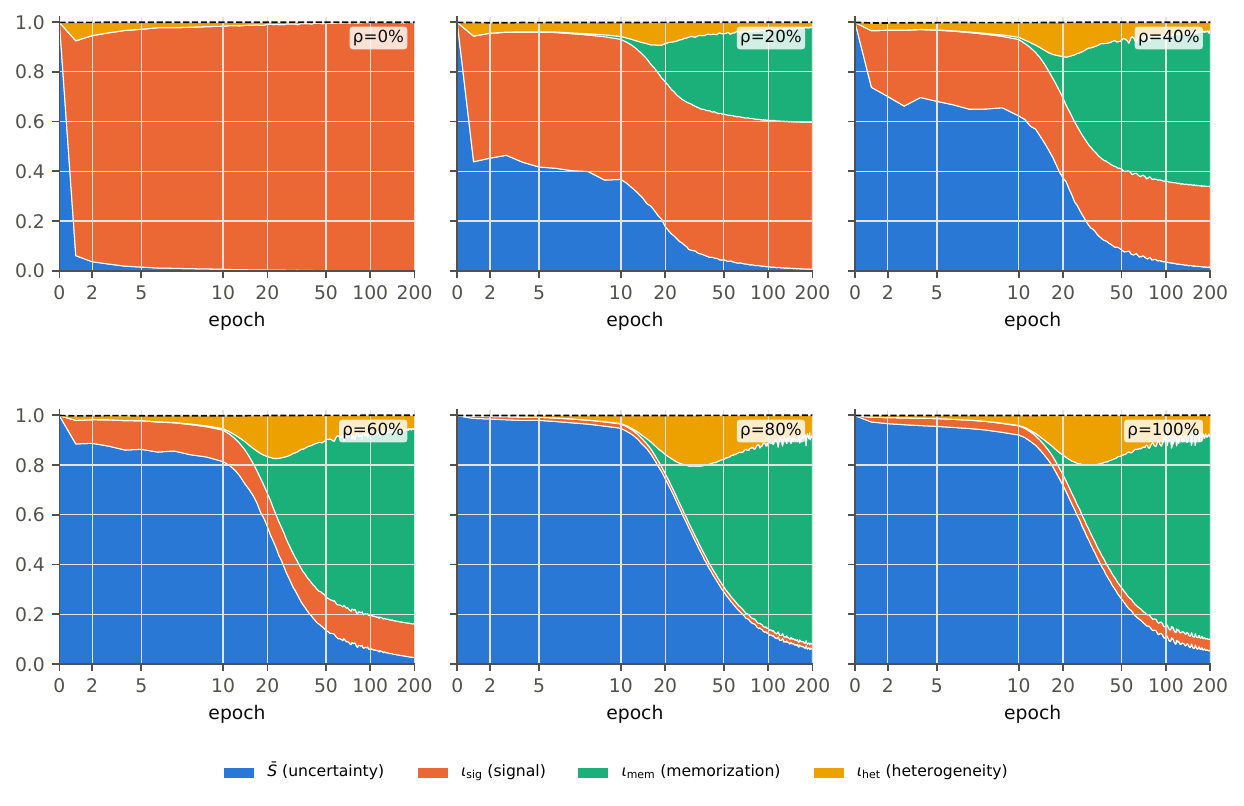}
\end{center}
\caption{Information budget (\ref{eq:budget}) on the training set under \textsf{UI} noise, averaged over 10 seeds. The stacked terms sum to $S(\pbar)$ (dashed).}
\label{fig:budget}
\end{figure}

The entropy of the mean prediction, $S(\pbar)$, never fell below 0.993 in any run or epoch of the main grid (Figure~\ref{fig:gap}, left). On these balanced datasets, it therefore carried little information about the regime. Instead, this information appeared in the gap $I(X;\Yh)/\ln N$. At the final epoch, the gap was 0.999 for clean labels and 0.993, 0.986, 0.974 and 0.943 for $\rho=0.2$-$0.8$. Because $S(\pbar)$ stayed this close to 1, the gap is approximately $1-\Sb$, the mean sharpness of the predictions, and the budget (Figure~\ref{fig:budget}) decomposes this sharpness. During early learning, the gap was dominated by $\iota_{\rm sig}$; during the memorization phase $\iota_{\rm mem}$ grew; and $\iota_{\rm het}$ was transiently large while the predictions within each $(y,\tilde y)$ cell were heterogeneous. The budget computed from argmax predictions, which is invariant to temperature, showed the same sequence of terms (Appendix, Figure~\ref{fig:hardbudget}). The maximum absolute residual of the identities was $1.2 \times 10^{-14}$ across all runs and epochs (H1), as expected of an implementation check.

\subsection{Null, agreement and end point of the memorization term}
\label{sec:valid}

\begin{figure}[h]
\begin{center}
\includegraphics[width=\textwidth]{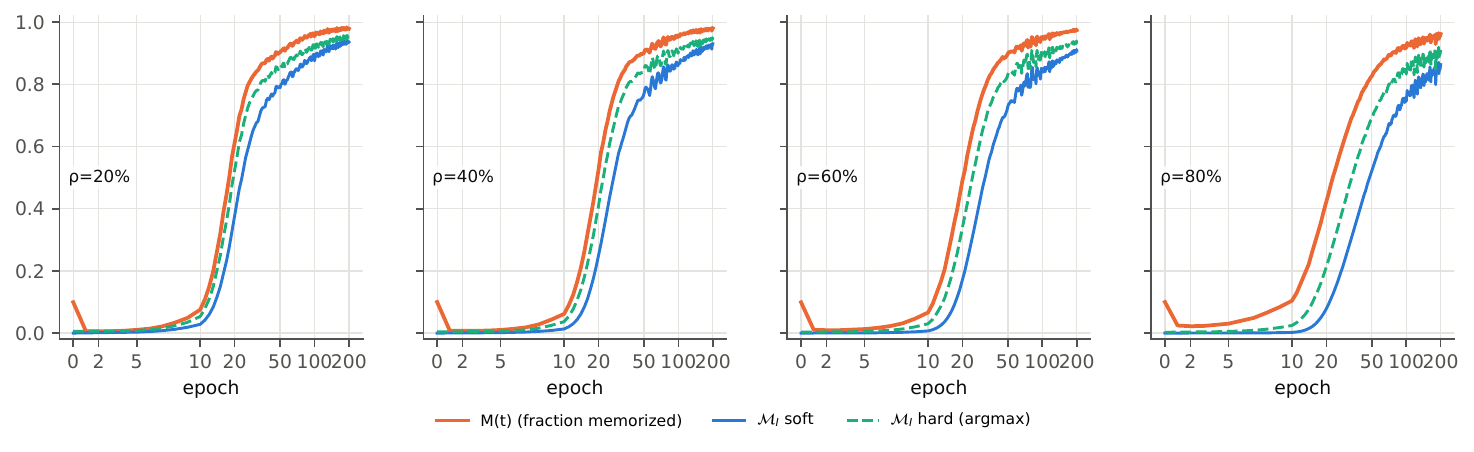}
\end{center}
\caption{Memorized fraction $M(t)$ and memorization information $\MI(t)$, computed from soft predictions and argmax (hard, temperature-invariant) predictions. \textsf{UI} noise, averaged over 10 seeds.}
\label{fig:estimands}
\end{figure}

\begin{figure}[h]
\begin{center}
\includegraphics[width=0.55\textwidth]{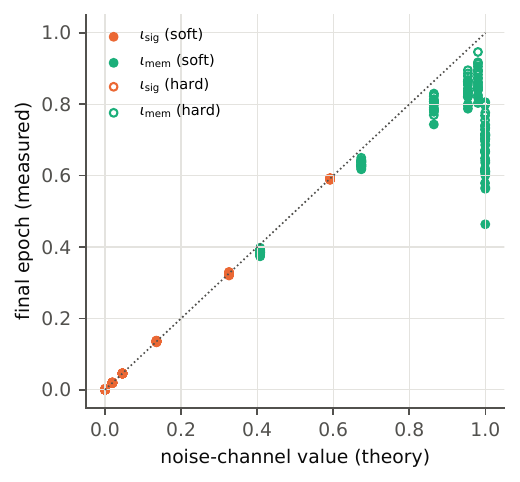}
\end{center}
\caption{Budget terms at epoch 200 compared with their complete-memorization values (Proposition~\ref{prop:endpoint}, computed from the empirical label channel), for all runs of the main grid and for both soft and hard budgets.}
\label{fig:endpoint}
\end{figure}

\begin{table}[t]
\caption{Budget terms at epoch 200 compared with their complete-memorization values, computed from the empirical label channel. \textsf{UI} noise, averaged over 10 seeds.}
\label{tab:endpoint}
\begin{center}
\small
\begin{tabular}{lcccccc}
\multicolumn{1}{c}{\bf }  &\multicolumn{3}{c}{\bf $\iota_{\rm sig}$}  &\multicolumn{3}{c}{\bf $\iota_{\rm mem}$}
\\ \hline \\
\multicolumn{1}{c}{\bf $\rho$}  &\multicolumn{1}{c}{\bf REFERENCE}  &\multicolumn{1}{c}{\bf SOFT}  &\multicolumn{1}{c}{\bf HARD}  &\multicolumn{1}{c}{\bf REFERENCE}  &\multicolumn{1}{c}{\bf SOFT}  &\multicolumn{1}{c}{\bf HARD}
\\ \hline \\
0.2 & 0.592 & 0.590 & 0.592 & 0.408 & 0.382 & 0.388\\
0.4 & 0.326 & 0.325 & 0.326 & 0.674 & 0.627 & 0.640\\
0.6 & 0.135 & 0.135 & 0.136 & 0.865 & 0.786 & 0.809\\
0.8 & 0.020 & 0.019 & 0.020 & 0.980 & 0.846 & 0.893\\
1.0 & 0.046 & 0.046 & 0.046 & 0.954 & 0.821 & 0.862\\
\hline
\end{tabular}
\end{center}
\end{table}

\paragraph{Null (H2).} On held-out data with freshly drawn noisy labels, $I(\Yh;\Yt\mid Y)$ remained within the 95th percentile of its permutation null (100 permutations, logged during training) in 93\% of scheduled epochs. Because these epochs are strongly correlated, this fraction is not an error rate. A more appropriate check counts rejections across runs, which use independent noise realizations, at a fixed epoch. Across the 70 corrupted runs of the main grid, with 1000 permutations and $p=(1+k)/(1+B)$, where $k$ counts the permutations whose statistic reaches the observed value, the test-set null was rejected at the 5\% level in an average of 6.8\% of runs over the 37 scheduled epochs. The rate ranged from 2.9\% to 11.4\% depending on the epoch, and only one epoch fell outside the central 95\% range of a binomial distribution with 70 runs. Taking runs as the unit, the 95\% bootstrap CI of this mean rate was 5.1 to 8.7\%, slightly above the nominal 5\% ($t$-test against 5\%, $p=0.06$). Because the test labels were never used in training, the null holds by construction here, so this analysis checks the calibration of the test rather than the validity of $\MI$ as a measure of memorization. On the test set, the null-corrected index $(I-\overline{I}_{\rm null})/H(\Yt\mid Y)$ remained within $\pm 0.0034$ in every run and epoch. The uncorrected held-out value $\MI^{\rm out}$ (soft) grew slowly as predictions sharpened, to 0.014-0.031 at epoch 200 under \textsf{UI} noise, consistent with the upward bias of the plug-in estimate, whereas the in-sample value $\MI^{\rm in}$ reached 0.86-0.94 (Section~\ref{sec:mterm}). On the training set, the statistic exceeded the 99th percentile in 99\% of the epochs after $t_{10}$. Since the network was trained on $\tilde y$, this is expected and is not evidence of power.

\paragraph{Agreement with $M$ (H3).} The median Spearman correlation between $M(t)$ and $\MI(t)$ over epochs was 0.993 (Figure~\ref{fig:estimands}). Because both curves rise monotonically after onset, this statistic is insensitive to delays and monotone distortions and should be interpreted as a weak criterion. The delays were not negligible. For $t_{50}$, hard $\MI$ lagged $M$ by 1.7 to 2.5 epochs for $\rho\le0.4$, by 4.1 epochs at $\rho=0.6$ and by 10.8 epochs at $\rho=0.8$. Soft $\MI$ lagged by up to 25 epochs (Table~\ref{tab:mi_vs_m}). Because both curves rise with time, $\MI$ is close to a monotone function of $M$ within a run (a per-run isotonic fit left a root mean square residual of at most 0.014), so these lags mostly reflect the different scales of the two quantities ($\MI$ is not a fraction of samples) rather than a different ordering of the epochs.

\paragraph{End point.} At epoch 200, the signal term was within 0.002 (soft) and 0.001 (hard) of its complete-memorization value on average over seeds, and within 0.006 in every run (Figure~\ref{fig:endpoint}, Table~\ref{tab:endpoint}). With noisy-label training accuracies ranging from 96\% to 99\%, this agreement is close to mechanical and checks Proposition~\ref{prop:endpoint} rather than adding evidence. The more informative quantity is the distance of the memorization term from its end point before interpolation. This distance increased with $\rho$ (soft: 0.026 at $\rho=0.2$, 0.134 at $\rho=0.8$), consistent with incomplete memorization ($M$ between 0.96 and 0.98 for \textsf{UI}; 0.86 for \textsf{SU} and 0.88 for \textsf{PERM} at $\rho=1$) and residual uncertainty in the soft predictions.

\subsection{Memorization information versus memorized fraction}
\label{sec:mi_vs_m}

\begin{figure}[h]
\begin{center}
\includegraphics[width=\textwidth]{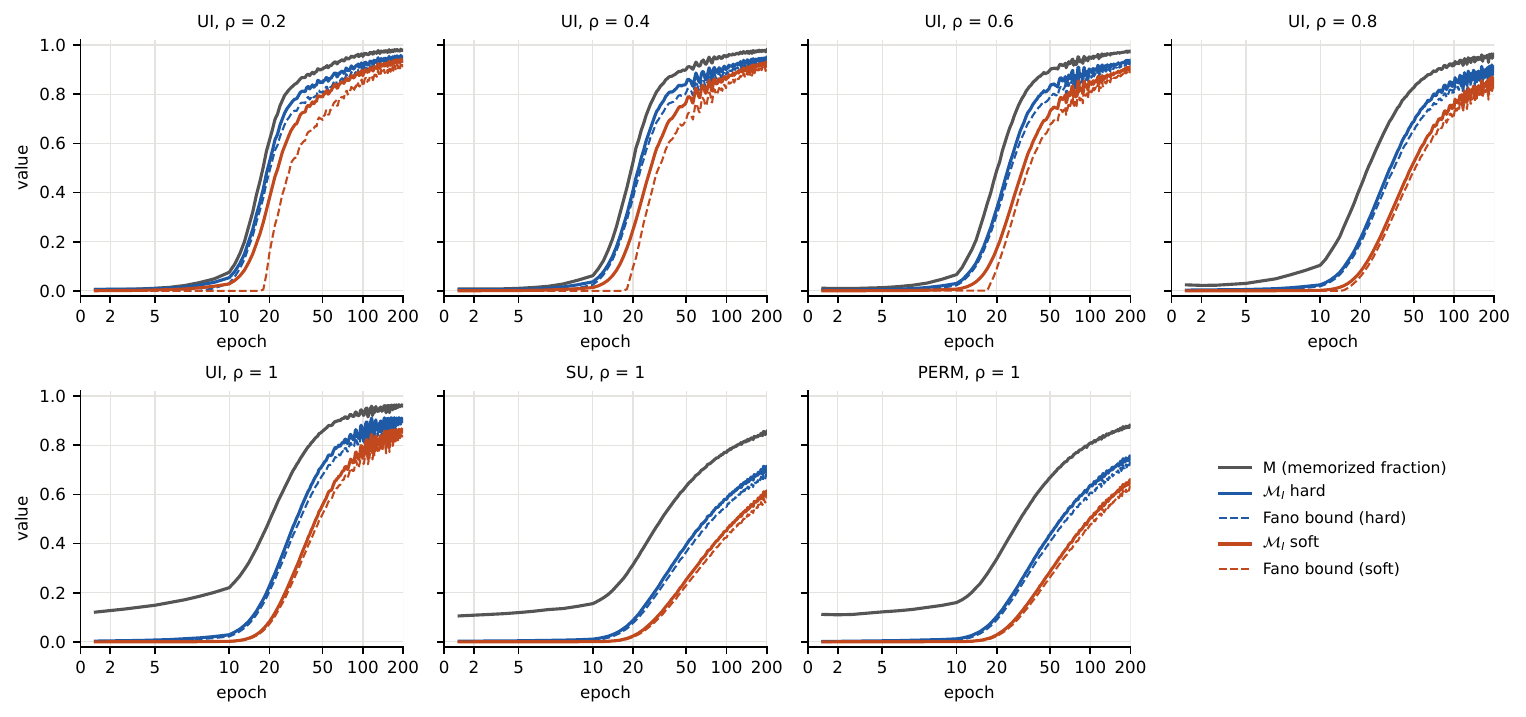}
\end{center}
\caption{$M$, the hard and soft $\MI$, and the Fano lower bounds of Proposition~\ref{prop:fano} (dashed, clipped at 0), main grid, mean over 10 seeds. Under \textsf{SU}, \textsf{PERM} and \textsf{UI} at $\rho=1$, $M$ starts near its chance level ($\approx0.1$) while $\MI$ starts near 0.}
\label{fig:fano}
\end{figure}

\paragraph{The Fano bound in practice (exploratory).} As an implementation check, Proposition~\ref{prop:fano} was evaluated at every epoch on the 70 corrupted runs of the main grid and the 40 pair-flip runs, with no violation of the global bound in either the hard or the soft version. The per-class refinement, computed with the estimator $g$ of Proposition~\ref{prop:fano}, was not violated either, and it was never weaker than the global bound. The bound was close to $\MI$. Averaged over all epochs, the distance between hard $\MI$ and its lower bound was 0.022-0.026 under \textsf{UI}, \textsf{SU} and \textsf{PERM} noise and 0.028-0.033 under pair-flip noise. This average includes early epochs when both quantities were near zero. During the memorization phase, defined by $t_{10}$ and $t_{90}$ of the hard $\MI$, the distance ranged from 0.032 to 0.040 under \textsf{UI} noise, 0.028 under \textsf{SU} and \textsf{PERM}, and 0.045-0.051 under pair-flip noise. For $0.3 \le M \le 0.7$, its value ranged from 0.024 to 0.039 (Figure~\ref{fig:fano}). In these regimes, therefore, $\MI$ was close to what the training accuracies $(A_{\mathcal L},M)$ imply, which is consistent with the high correlation in H3. What $\MI$ adds is a zero population null under Assumption~\ref{ass:A1}, hence no chance level, and sensitivity to corrupted samples for which neither the clean nor the noisy label is predicted. The fraction $o$ of these off-target predictions peaked within the memorization phase, reaching 0.36 around epoch 16 for \textsf{UI} with $\rho=0.8$, and fell to 0.03 by epoch 200.

\paragraph{The chance level of $M$.} If predictions do not depend on the samples that were corrupted, $\MI$ is zero in the population (Proposition~\ref{prop:null}), whereas $M$ equals its chance level $M_{\rm chance}$ (Definition~\ref{def:M}). Table~\ref{tab:chance} shows that, after one epoch, $M$ was close to and slightly above $M_{\rm chance}$ in every condition, while hard $\MI$ was near zero. The chance level was at most 0.021 for \textsf{UI} noise with $\rho \leq  0.8$. It ranged from 0.10 to 0.11 at $\rho=1$ and 0.19 under pair-flip noise with $\rho=0.4$, where the network initially predicted $y+1$ for part of class $y$, regardless of which samples were corrupted. 

Under these conditions, the pre-specified $t_{10}$ was reached within the first few epochs, even in the absence of memorization. It decreased from $9.2$ to $1.8$ epochs as the pair-flip rate increased and was less than one epoch at $\rho=1$. The chance-corrected $\kappa$ removed most of this offset: its $t_{10}$ ranged from $9.6$ to $13.5$ epochs at $\rho=1$ and from $6.9$ to $9.8$ epochs under pair-flip noise. Hard $\MI$ was somewhat more stable across conditions, with $t_{10}$ ranging from $11.0$ to $11.4$ epochs under pair-flip noise and from $15.3$ to $21.7$ epochs at $\rho=1$, although part of this difference reflects its scale. For \textsf{UI} noise with $\rho \le 0.8$, $M$, $\kappa$ and hard $\MI$ gave onset times within approximately 5 epochs of one another. The onset-based conclusions of Table~\ref{tab:dyn} therefore seem to hold in these conditions, whereas later times such as $t_{50}$ depend on the measure (Table~\ref{tab:mi_vs_m}). The main weakness of $M$ is its chance level, which the correction largely addresses. $\MI$ avoids this offset by construction and adds the information-theoretic interpretation of Proposition~\ref{prop:fano}. All three measures ($M$, $\kappa$ and $\MI$) require clean labels. 

\begin{table}[t]
\caption{\textit{Chance level of $M$ after one epoch}. Means over 10 seeds.  $M_{\rm chance}$ is the expected value of $M$ if predictions do not depend on which samples were corrupted (Definition~\ref{def:M}).}
\label{tab:chance}
\begin{center}
\begin{tabular}{llccc}
\multicolumn{1}{c}{\bf NOISE}  &\multicolumn{1}{c}{\bf $\rho$}  &\multicolumn{1}{c}{\bf $M$}  &\multicolumn{1}{c}{\bf $M_{\rm chance}$}  &\multicolumn{1}{c}{\bf HARD $\MI$}
\\ \hline \\
\textsf{UI} & 0.2 & 0.006 & 0.005 & 0.006\\
\textsf{UI} & 0.4 & 0.008 & 0.006 & 0.004\\
\textsf{UI} & 0.6 & 0.010 & 0.009 & 0.004\\
\textsf{UI} & 0.8 & 0.024 & 0.021 & 0.003\\
\textsf{UI} & 1 & 0.121 & 0.112 & 0.003\\
\textsf{SU} & 1 & 0.105 & 0.101 & 0.002\\
\textsf{PERM} & 1 & 0.112 & 0.109 & 0.001\\
PF & 0.1 & 0.008 & 0.006 & 0.003\\
PF & 0.2 & 0.011 & 0.009 & 0.002\\
PF & 0.3 & 0.044 & 0.036 & 0.002\\
PF & 0.4 & 0.210 & 0.194 & 0.003\\
\hline
\end{tabular}
\end{center}
\end{table}

\begin{table}[t]
\caption{\textit{Memorization times.} Times in epochs, averaged over 10 seeds, exploratory. Times are based on $M$ (pre-specified, after its minimum), chance-corrected $\kappa$, and hard and soft $\MI$ (without onset correction). Thresholds on $\MI$ apply to an information quantity, not a fraction of samples.}
\label{tab:mi_vs_m}
\begin{center}
\begin{tabular}{llcccccccc}
\multicolumn{1}{c}{\bf }  &\multicolumn{1}{c}{\bf }  &\multicolumn{4}{c}{\bf $t_{10}$}  &\multicolumn{4}{c}{\bf $t_{50}$}
\\ \cline{3-6} \cline{7-10} \\
\multicolumn{1}{c}{\bf NOISE}  &\multicolumn{1}{c}{\bf $\rho$}  &\multicolumn{1}{c}{\bf $M$}  &\multicolumn{1}{c}{\bf $\kappa$}  &\multicolumn{1}{c}{\bf $\MI$ HARD}  &\multicolumn{1}{c}{\bf $\MI$ SOFT}  &\multicolumn{1}{c}{\bf $M$}  &\multicolumn{1}{c}{\bf $\kappa$}  &\multicolumn{1}{c}{\bf $\MI$ HARD}  &\multicolumn{1}{c}{\bf $\MI$ SOFT}
\\ \hline \\
\textsf{UI} & 0.2 & 11.0 & 11.1 & 12.2 & 14.0 & 18.2 & 18.5 & 19.9 & 23.7\\
\textsf{UI} & 0.4 & 12.0 & 12.0 & 13.0 & 16.0 & 20.1 & 20.2 & 22.6 & 27.5\\
\textsf{UI} & 0.6 & 11.8 & 12.0 & 14.0 & 18.0 & 20.9 & 21.6 & 25.0 & 32.2\\
\textsf{UI} & 0.8 & 10.3 & 11.9 & 15.6 & 21.9 & 23.2 & 25.1 & 34.0 & 48.2\\
\textsf{UI} & 1 & 0.5 & 9.6 & 15.3 & 21.6 & 20.6 & 23.1 & 32.4 & 45.2\\
\textsf{SU} & 1 & 0.9 & 13.5 & 21.7 & 31.1 & 33.9 & 39.2 & 73.9 & 110.8\\
\textsf{PERM} & 1 & 0.6 & 12.8 & 20.5 & 29.0 & 30.7 & 35.1 & 61.8 & 99.0\\
PF & 0.1 & 9.2 & 9.7 & 11.1 & 13.5 & 17.4 & 18.7 & 21.4 & 25.6\\
PF & 0.2 & 8.7 & 9.8 & 11.4 & 14.9 & 16.9 & 18.3 & 22.0 & 27.0\\
PF & 0.3 & 6.5 & 9.0 & 11.4 & 15.7 & 15.9 & 18.4 & 22.2 & 28.2\\
PF & 0.4 & 1.8 & 6.9 & 11.0 & 16.0 & 12.2 & 17.4 & 22.7 & 28.5\\
\hline
\end{tabular}
\end{center}
\end{table}

\subsection{Noise models were not interchangeable at complete corruption}
\label{sec:noise}

\begin{figure}[h]
\begin{center}
\includegraphics[width=\textwidth]{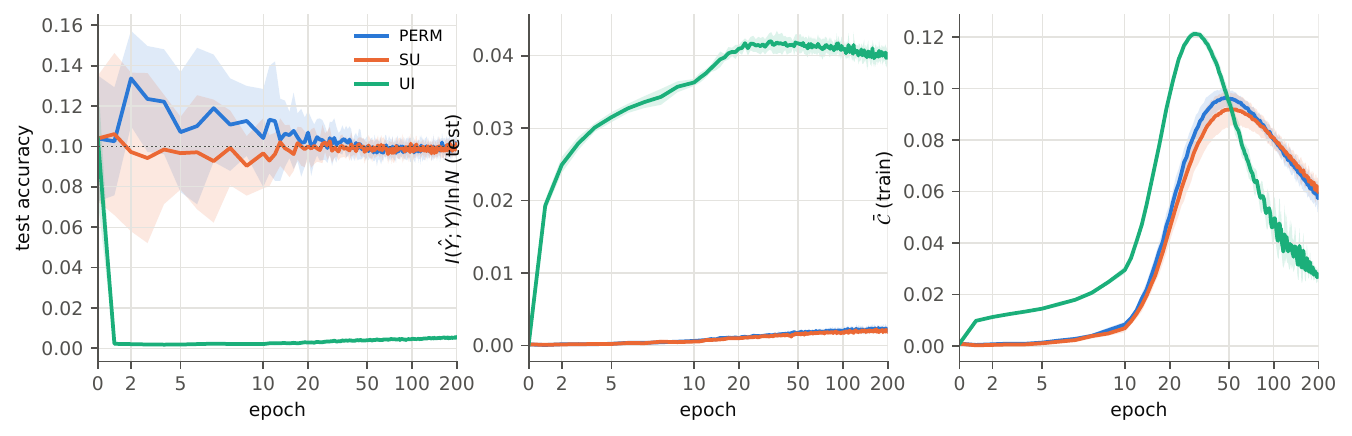}
\end{center}
\caption{All labels corrupted ($\rho=1$) under three mechanisms. Left: test accuracy (dotted line: chance). Center: information about the true class in the test predictions, $I(\Yh;Y)/\ln N$. Right: $\Cb$ on the training set. Mean $\pm$ standard deviation over 10 seeds.}
\label{fig:noise}
\end{figure}

The chance level of $M$ at $\rho=1$ already suggests differences among the three noise models. Under \textsf{UI} noise at $\rho=1$, every training label is a complementary label (``not this class''), and networks have been shown to learn from such labels~\citep{Ishida2017}. Final test accuracy was 0.5\% [0.5, 0.6], well below chance, and the test predictions carried $\iota_{\rm sig}=0.040$ [0.040, 0.041] of information about the true class (Figure~\ref{fig:noise}, H8). Under \textsf{SU} and \textsf{PERM}, test accuracy was at chance (10.1\%) and the information was 0.002. The two mechanisms that produce labels independent of the inputs were also harder to fit: 86\% (\textsf{SU}) and 88\% (\textsf{PERM}) of the training labels were memorized at epoch 200, compared with 96\% for \textsf{UI}. On Fashion-MNIST, \textsf{UI} noise at $\rho=1$ likewise yielded a test accuracy of 1.6\% (Section~\ref{sec:robust}). Studies that use full \textsf{UI} corruption as a substitute for random labels may therefore be studying a learnable task, which uniform resampling (\textsf{SU}) avoids.

\subsection{What the mean complexity tracked}
\label{sec:mechanism}

\begin{figure}[h]
\begin{center}
\includegraphics[width=\textwidth]{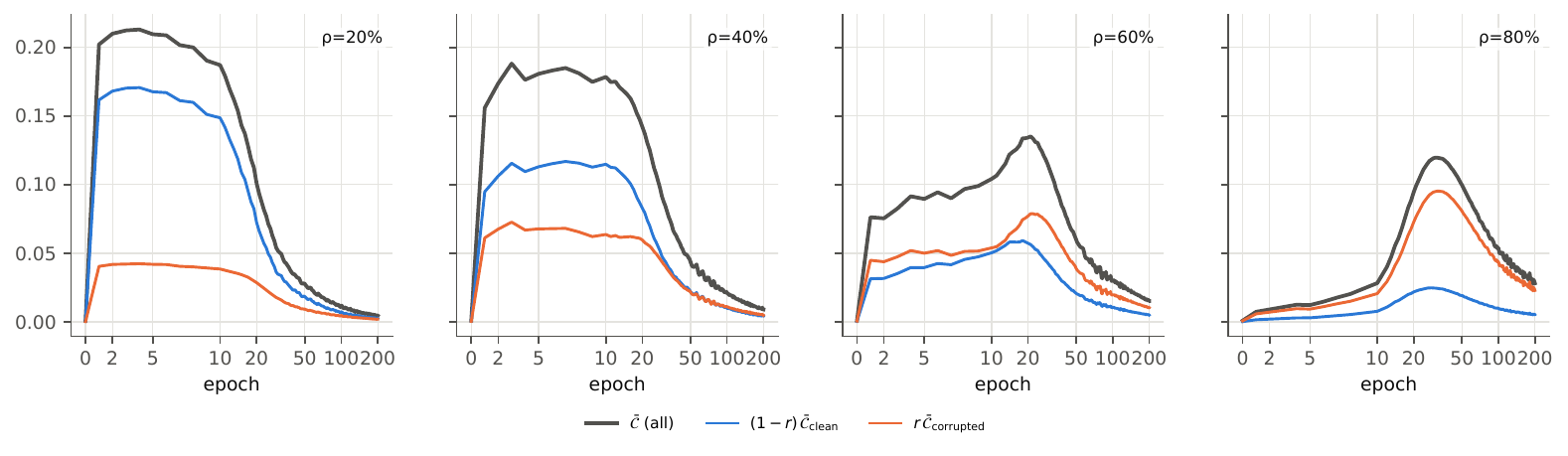}
\end{center}
\caption{\textit{Components of mean complexity}. Decomposition $\Cb=(1-r)\Cb_{\mathcal L}+r\,\Cb_{\mathcal N}$ (Observation~\ref{obs:mixture}) into the contributions of clean and corrupted samples. \textsf{UI} noise, averaged over 10 seeds.}
\label{fig:mixture}
\end{figure}

\begin{table}[t]
\caption{Peak times of $\Cb$ and its components, and training confidence at the peak of $\Cb$ (\textsf{UI} noise, means over 10 seeds; exploratory). Confidence is the mean of $\max_kp_k$ on the training set. At $\rho=0$ there are no corrupted samples, so $\Cb=\Cb_{\mathcal L}$. At $\rho=1$ there are no clean samples, so $\Cb=\Cb_{\mathcal N}$.}
\label{tab:mixture}
\begin{center}
\small
\begin{tabular}{lcccccc}  
\multicolumn{1}{c}{\bf $\rho$}  &\multicolumn{1}{c}{\bf $t^*(\Cb)$}  &\multicolumn{1}{c}{\bf $t^*(\Cb_{\mathcal N})$}  &\multicolumn{1}{c}{\bf $t^*(\Cb_{\mathcal L})$}  &\multicolumn{1}{c}{\bf SHARE OF $\Cb$ FROM CORRUPTED}  &\multicolumn{1}{c}{\bf CONF. AT $t^*(\Cb)$}  &\multicolumn{1}{c}{\bf $\Sb$ AT $t^*(\Cb)$}
\\ \hline \\
0 & 1.0 & -- & 1.0 & 0.00 & 0.958 & 0.061\\
0.2 & 3.7 & 3.8 & 3.7 & 0.20 & 0.737 & 0.465\\
0.4 & 5.7 & 5.0 & 6.2 & 0.38 & 0.607 & 0.629\\
0.6 & 19.8 & 22.0 & 16.3 & 0.57 & 0.583 & 0.551\\
0.8 & 31.8 & 33.1 & 28.4 & 0.80 & 0.607 & 0.498\\
1.0 & 29.7 & 29.7 & -- & 1.00 & 0.591 & 0.502\\
\hline
\end{tabular}
\end{center}
\end{table}

This section addresses the exploratory question of what $\Cb$ tracks (Section~\ref{sec:intro}).

\paragraph{Components.} The mean entropy decreased and the mean disequilibrium increased over training at every $\rho$, so that $\Cb$ rose and then fell, with a peak that occurred later at higher $\rho$ (Appendix, Figure~\ref{fig:components}). Two observations stand out in Table~\ref{tab:mixture} and Figure~\ref{fig:mixture}. First, both components of $\Cb$ peaked later as $\rho$ increased: $t^*(\Cb_{\mathcal N})$ shifted from  3.8 to 33.1 epochs and $t^*(\Cb_{\mathcal L})$ from 3.7 to 28.4. Clean samples therefore cannot simply be equated with ``learning'' and corrupted samples with ``memorization''. Second, at the peak of $\Cb$, the share contributed by corrupted samples was close to their fraction $r$, indicating similar mean complexity in the two groups at that epoch. Away from the peak the two groups differed: for $\rho \le 0.4$ the clean samples contributed most of $\Cb$ during early training, and for $\rho=0.8$ the corrupted samples did (Figure~\ref{fig:mixture}).

\paragraph{Timing relative to memorization and confidence.}
We compare the timing of the complexity peak with the progression of memorization and the rise in confidence. Complexity $\mathcal{C}(p)$ is a non-monotonic function of the confidence of $p$: it vanishes at the uniform distribution and at the vertices, and, for two-level distributions, reaches its maximum when the largest probability is approximately $0.76$. For $\rho\leq0.4$, the peak of $\Cb$ occurred at median epochs $4$ and $6.5$ across runs, before the onset of memorization, as measured by $t_{10}$ computed from $\kappa$, which ranged from $11$ to $12$ epochs. The peak therefore preceded measurable memorization in these conditions. For $\rho\geq0.6$, it occurred within the memorization phase. In these conditions, mean training confidence at the peak averaged between $0.58$ and $0.61$, with individual run values ranging from $0.51$ to $0.64$ when \textsf{SU} and \textsf{PERM} at $\rho=1$ were also included. The median distance between the peak epoch and the first epoch at which mean confidence reached $0.6$ was $1.5$ epochs at $\rho=0.6$ and $1$ epoch at $\rho=0.8$. At $\rho=0.6$, however, the peak was equally close to $t_{50}$ computed from $\kappa$, with a median distance of $1.5$ epochs; at $\rho=0.8$, the corresponding distance was $6.5$ epochs. Because the rise in confidence at high $\rho$ largely accompanies the fitting of corrupted samples, this design does not allow the timing associated with confidence to be separated from that associated with memorization. Across the 40 runs with $0<\rho<1$, the peak epoch and the confidence-crossing epoch had a Spearman correlation of $0.93$. This value, however, mostly reflected differences between noise rates: within individual noise rates, the correlation ranged from $0.24$ to $0.57$. With clean labels ($\rho=0$), $\Cb$ peaked at epoch $1$ in all ten runs, by which time mean training confidence had already reached values between $0.95$ and $0.96$ (Table~\ref{tab:mixture}). Logging once per epoch does not allow a crossing of the $0.6$ confidence level within the first epoch to be observed. The association between the complexity peak and a confidence of approximately $0.6$ can therefore be examined only where confidence rises slowly, which in these experiments occurred at $\rho\geq0.6$.

\paragraph{Two further checks.} Two further checks indicate that the mean confidence alone does not determine the peak. First, a model in which every sample has the same two-level distribution, with the observed mean confidence as its largest probability, predicts the peak when the mean confidence reaches 0.76. The predicted peak came later than the observed one: median epochs 22, 29 and 47 against 6.5, 20 and 31 at $\rho=0.4$, 0.6 and 0.8 (both 4 at $\rho=0.2$). Since $\Cb$ averages $\mathcal C(p_i)$ over samples with very different confidences, its maximum depends on the spread of the confidences, not only on their mean, which is consistent with a confidence of about 0.6, rather than 0.76, at the peak. Second, the median Spearman correlation between $\Cb$ and $\Sb$ over all epochs ranged from $0.99$ to $1.00$ for $\rho\leq0.6$, although this correlation was dominated by the decline after the peak. Over the first $2t^*(\Cb)+2$ epochs, the corresponding values were $0.73$, $0.04$, $0.08$, and $-0.36$ for $\rho=0.2$, $0.4$, $0.6$, and $0.8$, respectively. These findings suggest that the peak of $\Cb$ reflects changes in the spread of per-sample confidences during training. At high $\rho$, these changes coincided with memorization and could not be separated from it within our experimental design. At low $\rho$, the peak occurred before memorization began.

\paragraph{Trends with $\rho$ and the peak of $V$ (H4).} Across the 40 runs with $\rho\in\{0.2,\dots,0.8\}$, the peak of $\Cb$ occurred later as $\rho$ increased (Spearman correlation 0.93), whereas the best test epoch occurred slightly earlier (Spearman $-0.67$). The relation between $t^*(\Cb)$ and $\rho$ is step-like (Table~\ref{tab:mixture}), so a linear slope provides only a coarse summary. As a coarse summary, OLS gave slopes of $+49.2$ epochs per unit of $\rho$ for $t^*(\Cb)$ and $-4.2$ for the best test epoch; given the step-like relation, we do not interpret their confidence intervals. The peak of $V$ behaved similarly to that of $\Cb$ (Spearman correlation 0.97; Appendix, Figure~\ref{fig:variance}) and fell within $[t_{10},t_{90}]$ in all 40 runs (H4, first part). This window is wide (ranging from 10 to 72 epochs), so the result carries little weight. The second part of H4 failed: the peak of the label-free heterogeneity term $I(\Yh;X\mid\Yt)$ fell within the memorization phase in only 55\% of runs, so this term did not single out that phase. Defining the phase through $\kappa$ or through hard $\MI$ instead of $M$ left both results unchanged: the peak of $V$ fell within the phase in all 40 runs, and that of $I(\Yh;X\mid\Yt)$ in 55\% ($\kappa$) and 53\% (hard $\MI$) of runs. In the nine crossed runs at $\rho=0.4$, the variation in $t^*(\Cb)$ appeared to be associated more with the noise realization than with the initialization; with nine runs, this observation is descriptive only.

\paragraph{Sensitivity (H9) and stability (H10).} The peak of $\Cb$ was also sensitive to logit scaling (H9). Pooled over $\rho$, Kendall's $\tau$ between peak times obtained with $\mathcal D_2$ and its variants was 0.80 ($\mathcal D_1$), 0.84 ($\mathcal D_{JS}$), 0.46 (temperature 0.5) and 0.75 (temperature 2). Pooling inflates $\tau$ through the common trend with $\rho$. Within each $\rho$, $\tau$ ranged from $-0.58$ to $1.0$, and the median absolute shift of the peak time ranged from 0 to 10 epochs across functionals and from 1 to 21.5 epochs across temperatures, reinforcing the negative outcome of H9. Finally, the centroid trajectories $(\Sb(t),\Db_2(t))$ of all corruption rates remained close to the upper boundary of the admissible region and to one another (Appendix, Figure~\ref{fig:plane}). Their seed-to-seed variation, measured by the discrete Fr\'echet distance between trajectories, exceeded the separation between some corruption rates (H10, ratio of the largest mean within-rate distance to the smallest mean between-rate distance $R=2.30$).

\subsection{Complexity-based diagnostics as tools}
\label{sec:tools}

\begin{figure}[h]
\begin{center}
\includegraphics[width=0.75\textwidth]{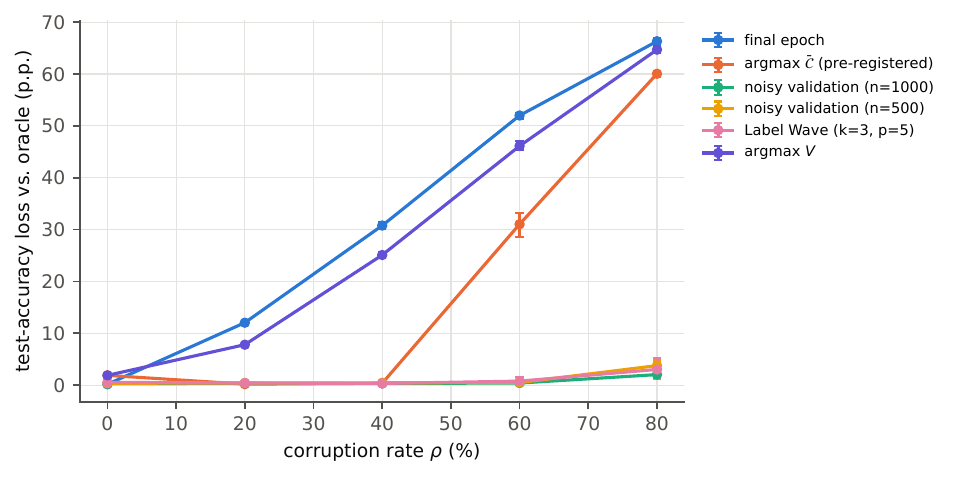}\\[4pt]
\includegraphics[width=\textwidth]{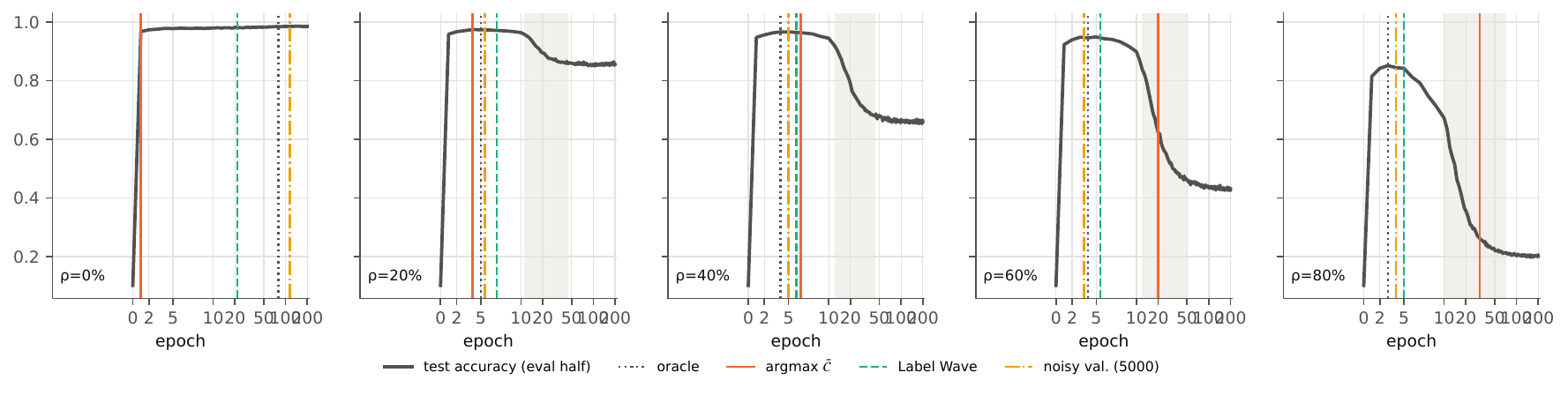}
\end{center}
\caption{\textit{Early-stopping benchmark}. Top: loss in test accuracy for each stopping rule versus the corruption rate (mean and 95\% CI). Bottom: median stopping epoch of each rule, shown on the mean test-accuracy curve for the evaluation half. Shading indicates the memorization phase defined using $M$.}
\label{fig:es}
\end{figure}

\paragraph{Early stopping.} Against the pre-specified baselines, stopping at the peak of $\Cb$ reduced the loss of test accuracy by 16.3 percentage points on average (H5), evaluated on the full test set as pre-specified. All 40 paired differences were negative, so the Holm-adjusted $p=1.8\times10^{-12}$ is the smallest value the one-sided test can attain with 40 runs ($2 \times 2^{-40}$). The gain over the loss plateau varied widely with $\rho$: 11.6, 28.5, 19.4 and 5.7 points for $\rho=0.2$, 0.4, 0.6 and 0.8 (on the evaluation half used in Table~\ref{tab:es}, the corresponding values are 11.9, 28.5, 19.2 and 5.1). At each rate, the gain was positive in all ten seeds (paired Wilcoxon test over independent runs, one-sided $p=0.001$, and $p=0.004$ after Holm adjustment over the four rates), so the conclusion of H5 does not depend on the pooling discussed in Section~\ref{sec:protocol}. The comparison changed when methods designed for early stopping were included (Table~\ref{tab:es}, Figure~\ref{fig:es}). For $\rho\leq0.4$, stopping at the peak of $\Cb$ incurred a loss ranging from $0.2$ to $0.4$ percentage points, comparable to noisy validation and Label Wave, with overlapping intervals. At $\rho=0.6$ and $\rho=0.8$, however, the corresponding losses were $31.0$ and $60.0$ percentage points, respectively, whereas losses for noisy validation and Label Wave were at most a few percentage points. In paired tests conducted separately at each noise rate, noisy validation with 1000 samples and Label Wave ($k=1,p=3$) outperformed stopping at the peak of $\Cb$ at $\rho=0.6$ and $\rho=0.8$, with a Holm-adjusted one-sided $p$-value of $0.004$ for each comparison. For $\rho\leq0.4$, the differences were not statistically significant, with adjusted $p$-values of at least $0.10$. On clean data, stopping at the peak of $\Cb$ incurred a loss of $1.9$ percentage points because it stopped too early. Averaged over $\rho\in\{0.2,\dots,0.8\}$, noisy validation with 500 samples reduced the loss relative to stopping at the peak of $\Cb$ by $21.6$ percentage points [14.5, 29.1]. For Label Wave ($k=3,p=5$), the reduction was $21.8$ percentage points [14.7, 29.4]. When the 16 label-free candidates were ranked using the validation half and five seeds, the three Label Wave configurations with $k=1$ tied. The first of these, Label Wave ($k=1,p=3$), was selected. On the evaluation half, using the other five seeds and averaging over $\rho\in\{0.2,\dots,0.8\}$, the selected rule incurred a loss of $0.5$ percentage points [0.2, 1.1]. The corresponding loss was $22.2$ percentage points [11.9, 33.0] for stopping at the peak of $\Cb$ and $0.8$ percentage points [0.5, 1.2] for noisy validation with 1000 samples. The same rule had also been selected when selection and evaluation used the same test images. As noted in Section~\ref{sec:protocol}, rule selection used the clean labels of the validation half. For $0<\rho<1$, the patience $p$ of Label Wave did not change its stopping epoch: configurations with the same window $k$ yielded identical losses (Appendix, Table~\ref{tab:lw}). The signal that Label Wave tracks, the fraction of training samples whose prediction changes between epochs, had a minimum a few epochs after initialization and a second rise during memorization. Its stopping epoch fell near the first minimum, close to the best test epoch (Appendix, Figure~\ref{fig:lwsignal}). The loss of noisy validation decreased with the validation size mainly at $\rho = 0.8$ (3.8, 2.0 and 0.8 points with 500, 1000 and 5000 samples), and at $\rho \ge 0.6$ it was smaller than that of the peak of $\Cb$ for every size we tried (Appendix, Figure~\ref{fig:valsize}, left). The seed-split selection is shown in the same figure (right). We did not include progressive early stopping~\citep{Bai2021}, which requires a different training procedure.

\begin{table}[t]
\caption{\textit{Early-stopping evaluation}. Loss in clean test accuracy relative to the best epoch, evaluated on a held-out half of the test set under \textsf{UI} noise. Values are percentage points, reported as means and 95\% bootstrap CI over 10 seeds. All nine Label Wave configurations are reported in Table~\ref{tab:lw}.}
\label{tab:es}  
\begin{center}
\small
\setlength{\tabcolsep}{4pt}  
\begin{tabular}{lccccc}
\multicolumn{1}{c}{\bf RULE}  &\multicolumn{1}{c}{\bf $\rho=0$}  &\multicolumn{1}{c}{\bf $\rho=0.2$}  &\multicolumn{1}{c}{\bf $\rho=0.4$}  &\multicolumn{1}{c}{\bf $\rho=0.6$}  &\multicolumn{1}{c}{\bf $\rho=0.8$}
\\ \hline \\
Final epoch & 0.2 [0.1, 0.2] & 12.0 [11.6, 12.5] & 30.8 [30.0, 31.4] & 52.0 [51.4, 52.5] & 66.3 [65.7, 66.9]\\
Loss plateau & 0.8 [0.6, 1.0] & 12.1 [11.5, 12.6] & 28.9 [28.2, 29.6] & 50.2 [49.0, 51.4] & 65.1 [64.7, 65.6]\\
$\arg\max\Cb$ (pre-specified) & 1.9 [1.7, 2.0] & 0.2 [0.1, 0.3] & 0.4 [0.2, 0.6] & 31.0 [28.6, 33.2] & 60.0 [59.4, 60.7]\\
$\arg\max V$ & 1.9 [1.7, 2.0] & 7.8 [7.3, 8.3] & 25.1 [24.5, 25.8] & 46.1 [45.3, 47.1] & 64.7 [64.1, 65.5]\\
Noisy validation, 5000 & 0.2 [0.1, 0.3] & 0.4 [0.2, 0.6] & 0.2 [0.1, 0.4] & 0.2 [0.0, 0.4] & 0.8 [0.2, 1.5]\\
Noisy validation, 1000 & 0.3 [0.3, 0.3] & 0.4 [0.3, 0.4] & 0.3 [0.2, 0.4] & 0.4 [0.3, 0.6] & 2.0 [1.6, 2.6]\\
Noisy validation, 500 & 0.4 [0.3, 0.4] & 0.4 [0.3, 0.5] & 0.4 [0.4, 0.6] & 0.5 [0.4, 0.6] & 3.8 [2.8, 4.8]\\
Label Wave ($k=1,p=3$) & 0.5 [0.4, 0.6] & 0.2 [0.2, 0.3] & 0.1 [0.0, 0.3] & 0.2 [0.0, 0.3] & 1.1 [0.4, 2.0]\\
Label Wave ($k=3,p=5$) & 0.5 [0.3, 0.7] & 0.4 [0.3, 0.5] & 0.3 [0.2, 0.5] & 0.8 [0.2, 1.5] & 3.0 [1.2, 5.4]\\
\hline
\end{tabular}
\end{center}
\end{table}

\begin{figure}[h]
\begin{center}
\includegraphics[width=0.9\textwidth]{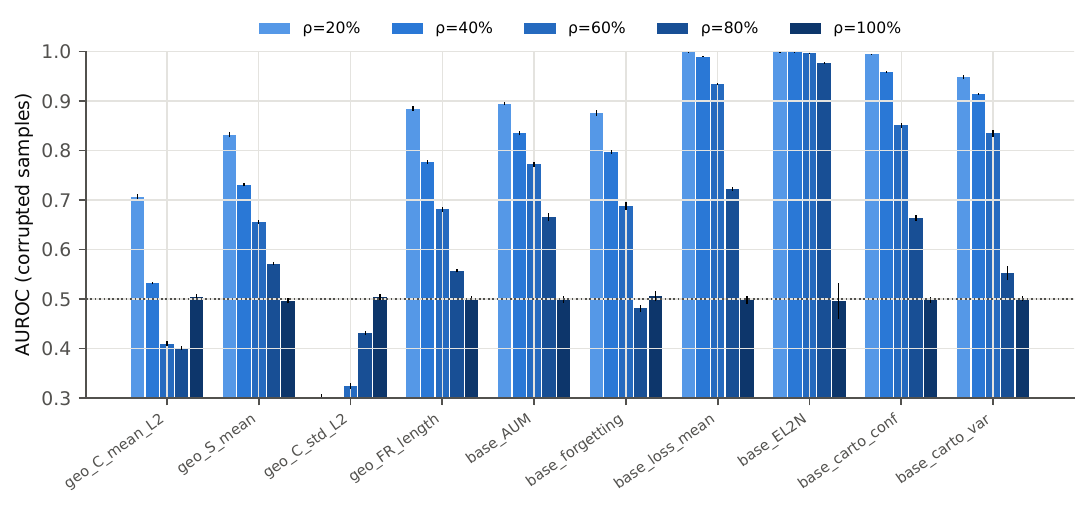}
\end{center}
\caption{\textit{Detection of corrupted samples}. AUROC, with the direction fixed in advance, for identifying corrupted training samples from features computed over the first 50 epochs (mean $\pm$ s.d.\ over 10 seeds). Geometric features (prefix \texttt{geo}: mean and standard deviation of $\mathcal C_i$ over epochs, mean of $S_i$ and Fisher-Rao path length) depend only on $p_i$. The baselines use the noisy label: area under the margin~\citep{Pleiss2020}, forgetting events~\citep{Toneva2019}, mean loss, EL2N~\citep{Paul2021} and dataset-cartography confidence and variability~\citep{Swayamdipta2020}.}
\label{fig:auroc}
\end{figure}

\paragraph{Per-sample detection.}
We evaluated the time average of $\mathcal C_i$ over the first 50 epochs as a score for detecting corrupted samples (H6 and Figure~\ref{fig:auroc}). With the score direction fixed in advance, its AUROC decreased from $0.707$ at $\rho=0.2$ to $0.401$ at $\rho=0.8$. An AUROC below $0.5$ indicates that the score tends to rank samples in the direction opposite to the one specified. 

The comparison with label-dependent baselines involves a difference in the information available to each method. While $\mathcal C_i$ and the other geometric features depend only on $p_i$, EL2N~\citep{Paul2021}, mean loss, area under the margin~\citep{Pleiss2020}, forgetting events~\citep{Toneva2019}, and cartography statistics~\citep{Swayamdipta2020} also use the noisy label $\tilde y_i$. The best-performing baseline, EL2N, achieved AUROC values between $0.977$ and $0.999$ when computed over the first 10 epochs, following \citet{Paul2021}. This difference in available information should be considered when interpreting the pre-specified non-inferiority comparison. Across the 60 runs included in the pre-specified analysis, the aggregate difference was $\Delta\mathrm{AUROC}=-0.318$. This analysis included \textsf{SU} and \textsf{PERM} at $\rho=1$, for which both scores performed at chance level. When the comparison was restricted to \textsf{UI} noise with $0<\rho<1$, the difference was $\Delta\mathrm{AUROC}=-0.481$.

A comparison with more closely matched information requirements is provided by simple label-free scores computed from the same
predictions (Table~\ref{tab:free}). These comprise entropy, the largest predicted probability, and the margin between the two largest probabilities, each averaged over epochs; the temporal variance of $p_t(k\mid x_i)$, averaged over classes; and the fraction of consecutive epochs in which the predicted class changed. Over the first 50 epochs, temporal variance achieved the highest AUROC among these scores at $\rho=0.2$, with a value of $0.969$, compared with $0.707$ for the time-averaged complexity. At $\rho=0.8$, the prediction-change score performed best, with an AUROC of $0.659$, compared with $0.401$ for time-averaged complexity. With the pre-specified direction, time-averaged complexity had an AUROC below $0.5$ in all ten seeds for $\rho\geq0.6$ over the 50-epoch window, and for $\rho\geq0.4$ over the 10 and 25-epoch windows. These results suggest that its usefulness as a ranking score depended in part on the noise rate and the observation window. Allowing the direction to be chosen retrospectively, as in the symmetric AUROC, resulted in values ranging from $0.53$ to $0.71$. Even with this adjustment, complexity remained below the best label-free score for every window and noise rate considered. The differences were small, however, for the 10-epoch window at $\rho\geq0.6$, ranging from $0.002$ to $0.004$.

For reference, the loss computed with the noisy labels achieved AUROC values between $0.977$ and $0.999$ over the first 10 epochs, when little memorization had occurred. Its discrimination was weaker over the 50-epoch window, with an AUROC of $0.722$ at $\rho=0.8$. Over the first 10 epochs, EL2N and noisy-label loss produced nearly identical rankings, with absolute per-run AUROC differences no greater than $0.001$. This similarity is consistent with both scores depending primarily on the probability assigned to the noisy label. Among the geometric features we computed, the best was the Fisher-Rao path length, defined as the sum over epochs of $2 \arccos \sum_k \sqrt{p_t(k)\,p_{t-1}(k)}$~\citep{Amari2016}, with AUROC from 0.884 at $\rho=0.2$ to 0.557 at $\rho=0.8$ (first 50 epochs). Temporal variance and prediction changes did better.

\begin{table}[t]
\caption{Label-free detection of corrupted samples (exploratory): AUROC with the direction fixed in advance, for scores averaged over the first $W$ epochs (\textsf{UI} noise; means over 10 seeds, standard deviations at most 0.014). The last row uses the noisy label and is shown for reference. With $W=50$ the logits are available at 40 epochs (every second epoch after epoch 30).}
\label{tab:free}
\begin{center}
\footnotesize
\setlength{\tabcolsep}{3pt}
\begin{tabular}{lcccccccccccc}
\multicolumn{1}{c}{\bf }  &\multicolumn{4}{c}{\bf $W=10$}  &\multicolumn{4}{c}{\bf $W=25$}  &\multicolumn{4}{c}{\bf $W=50$}
\\ \cline{2-5} \cline{6-9} \cline{10-13} \\
\multicolumn{1}{c}{\bf SCORE}  &\multicolumn{1}{c}{\bf $\rho=0.2$}  &\multicolumn{1}{c}{\bf $0.4$}  &\multicolumn{1}{c}{\bf $0.6$}  &\multicolumn{1}{c}{\bf $0.8$}  &\multicolumn{1}{c}{\bf $\rho=0.2$}  &\multicolumn{1}{c}{\bf $0.4$}  &\multicolumn{1}{c}{\bf $0.6$}  &\multicolumn{1}{c}{\bf $0.8$}  &\multicolumn{1}{c}{\bf $\rho=0.2$}  &\multicolumn{1}{c}{\bf $0.4$}  &\multicolumn{1}{c}{\bf $0.6$}  &\multicolumn{1}{c}{\bf $0.8$}
\\ \hline \\
entropy & 0.693 & 0.662 & 0.642 & 0.598 & 0.838 & 0.751 & 0.679 & 0.588 & 0.831 & 0.730 & 0.656 & 0.571\\
largest probability & 0.690 & 0.659 & 0.642 & 0.605 & 0.883 & 0.805 & 0.726 & 0.612 & 0.889 & 0.786 & 0.697 & 0.587\\
top-2 margin & 0.687 & 0.656 & 0.639 & 0.604 & 0.902 & 0.830 & 0.746 & 0.620 & 0.913 & 0.808 & 0.708 & 0.588\\
temporal variance & 0.619 & 0.518 & 0.445 & 0.438 & 0.905 & 0.799 & 0.630 & 0.480 & 0.969 & 0.936 & 0.875 & 0.589\\
prediction changes & 0.554 & 0.549 & 0.556 & 0.572 & 0.884 & 0.831 & 0.775 & 0.657 & 0.928 & 0.876 & 0.806 & 0.659\\
$\mathcal C_i$ (mean) & 0.531 & 0.366 & 0.360 & 0.399 & 0.665 & 0.419 & 0.298 & 0.387 & 0.707 & 0.532 & 0.410 & 0.401\\
noisy-label loss & 0.999 & 0.999 & 0.997 & 0.977 & 0.999 & 0.996 & 0.980 & 0.846 & 0.998 & 0.990 & 0.934 & 0.722\\
\hline
\end{tabular}
\end{center}
\end{table}

\subsection{Robustness across datasets, noise models and optimizers (exploratory)}
\label{sec:robust}

The results above come from one dataset, a family of symmetric noise models, and one optimizer. The 280 extension runs vary these factors in turn while retaining the architecture. Table~\ref{tab:robust} summarizes the results alongside those of the main grid. 

\begin{figure}[h!]
\begin{center}
\includegraphics[width=0.6\textwidth]{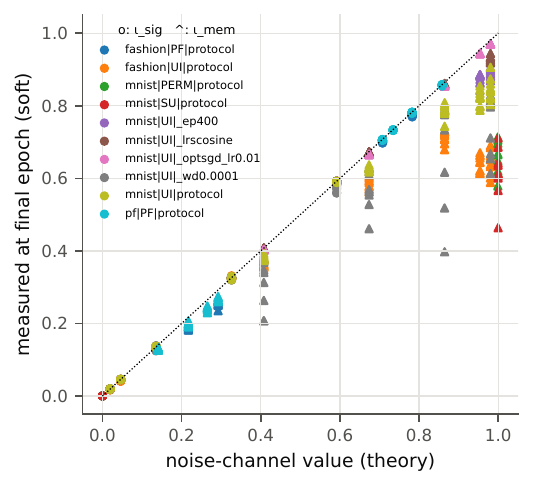}
\end{center}
\caption{\textit{End-point robustness}. Budget terms at the final epoch compared with their complete-memorization values across all conditions (circles: $\iota_{\rm sig}$; triangles: $\iota_{\rm mem}$, soft). The $\iota_{\rm mem}$ points on the diagonal at high noise correspond to runs that reached interpolation (SGD and cosine schedule).}
\label{fig:robust_endpoint}
\end{figure}

\paragraph{Measurement.} In every condition, the identities held with residuals $\le1.2\times10^{-14}$. The fraction of scheduled test epochs within the 95th percentile of the null ranged from 89\% to 97\% across conditions, a descriptive result subject to the caveats in Section~\ref{sec:valid}. The median Spearman correlation between $M$ and $\MI$ ranged from 0.977 to 1.000 across conditions with $0<\rho<1$ (soft and hard versions). At the final epoch, the signal term deviated from its reference value by at most 0.014 in any run, except under weight decay (0.032 at $\rho=0.2$), where predictions remained softer (Figure~\ref{fig:robust_endpoint}). 

SGD  reached interpolation in all its runs (median epoch 113-122 for $\rho>0$), and the cosine schedule reached interpolation  at $\rho\le0.6$. In those runs, the memorization term also reached its reference value: for example, 0.408 against 0.408 (cosine, $\rho=0.2$, soft), and 0.970 (soft) and 0.980 (hard) against 0.980 (SGD, $\rho=0.8$). As with the signal term, this is a sanity check of Proposition~\ref{prop:endpoint}. Training with Adam for 400 epochs did not lead to interpolation at $\rho\in\{0.8,1\}$: $M$ reached 0.975 and 0.978, and the soft memorization term reached  0.893 and 0.874, compared with 0.846 and 0.821 at epoch 200. The first 200 epochs of these runs were identical to the corresponding protocol runs, as expected under deterministic training.

\begin{figure}[h]
\begin{center}
\includegraphics[width=\textwidth]{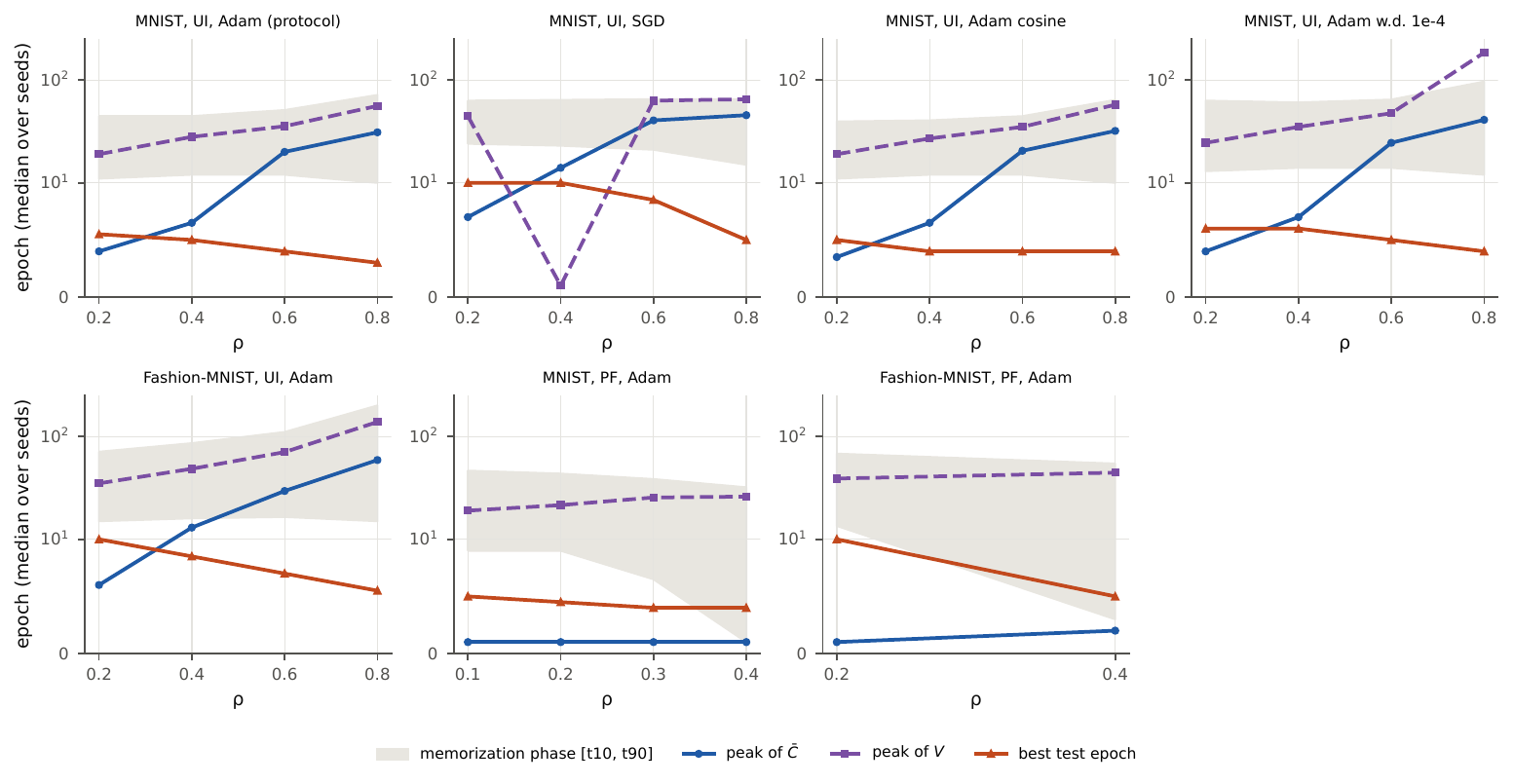}
\end{center}
\caption{\textit{Timing across conditions}. Peak of $\Cb$ ($t^*_{\bar C}$), peak of $V$ ($t^*_V$) and best test epoch versus $\rho$, by condition (medians over seeds). Shading indicates the memorization phase defined using $M$ under the pre-specified definition. Under pair-flip noise, its lower boundary is affected by the chance level of $M$ (Section~\ref{sec:mi_vs_m}). For SGD at $\rho=0.4$ the peak of $V$ is bimodal across runs (epoch 1 in six runs, 53-60 in four).}
\label{fig:robust_timing}
\end{figure}

\paragraph{Complexity.} The decomposition of Observation~\ref{obs:mixture} held to $10^{-16}$, and the share of $\Cb$ contributed by corrupted samples at its peak remained close to the corrupted fraction (0.20, 0.38, 0.57 and 0.79 on Fashion-MNIST). Under \textsf{UI} noise, the peak of $\Cb$ occurred later as $\rho$ increased, whereas the best test epoch  occurred earlier, in every training variant and on Fashion-MNIST (Table~\ref{tab:robust}, Figure~\ref{fig:robust_timing}). Under pair-flip noise, where at most 40\% of labels were corrupted and confidence rose quickly, the peak occurred at epoch 1 (median) at every rate on MNIST and at epochs 1 and 2 on Fashion-MNIST, so its correlation with  $\rho$ was weak ($r_s=-0.22$ and $0.27$, Table~\ref{tab:robust}). The trend of the peak with $\rho$ observed under \textsf{UI} noise therefore did not carry over to pair-flip noise. The peak of $V$ was less regular: with SGD at $\rho=0.4$, it occurred at epoch 1 in six runs and at epochs 53-60 in the other four, so its median in Figure~\ref{fig:robust_timing} falls at epoch 1.

\begin{figure}[h]
\begin{center}
\includegraphics[width=\textwidth]{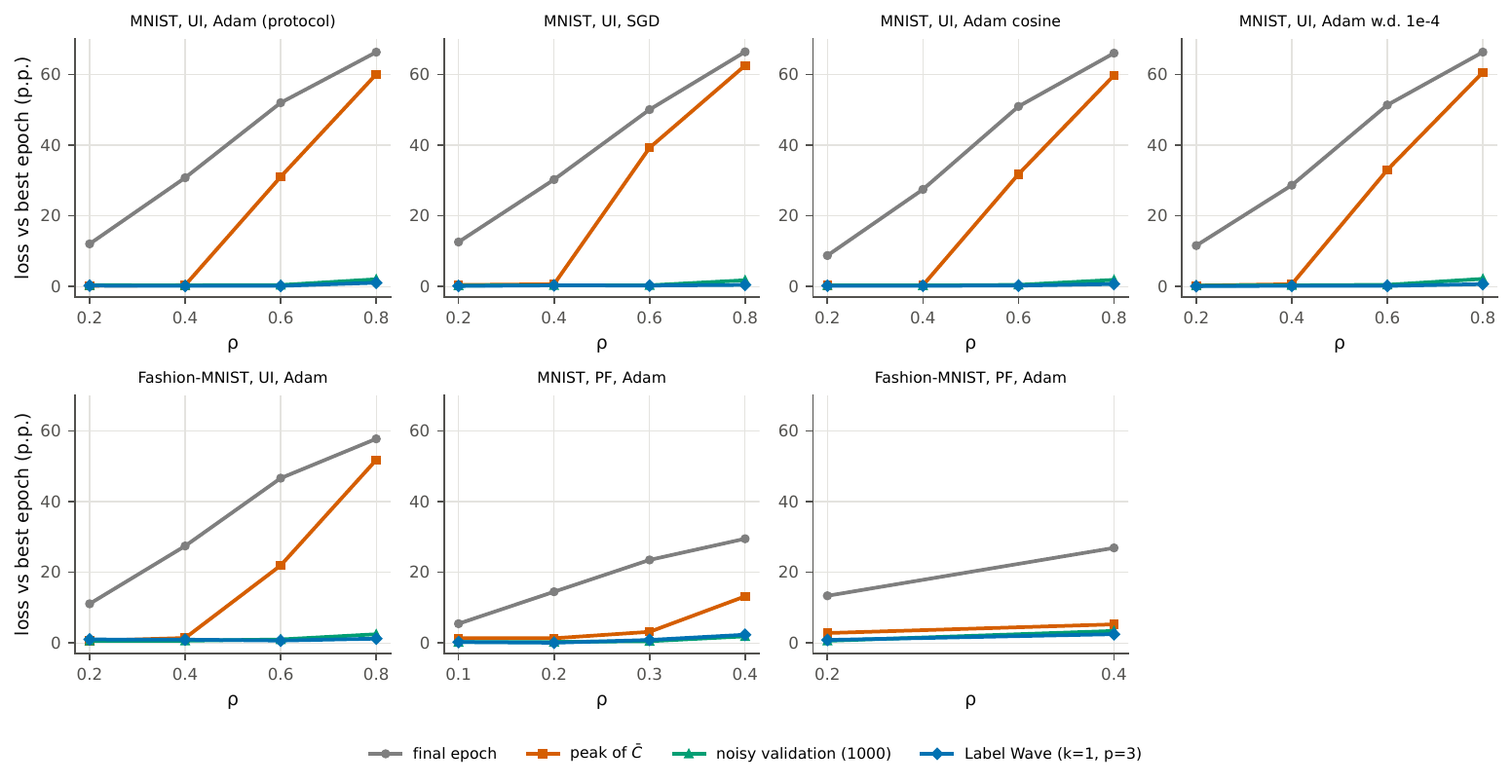}
\end{center}
\caption{\textit{Early stopping across conditions}. Loss in clean test accuracy, in percentage points, for four stopping rules versus $\rho$, by  condition (means over seeds).}
\label{fig:robust_es}
\end{figure}

\paragraph{Early stopping.} The ranking observed on the main grid held across all conditions (Table~\ref{tab:robust}, Figure~\ref{fig:robust_es}). Averaged over $0<\rho<1$, noisy validation with 1000 samples and Label Wave incurred losses of 0.3-2.0 percentage points. Stopping at the peak of $\Cb$ incurred losses of 19-26 percentage points under \textsf{UI} noise, mostly at high $\rho$, and 4-5 percentage points under pair-flip noise.

\begin{figure}[h]
\begin{center}
\includegraphics[width=0.75\textwidth]{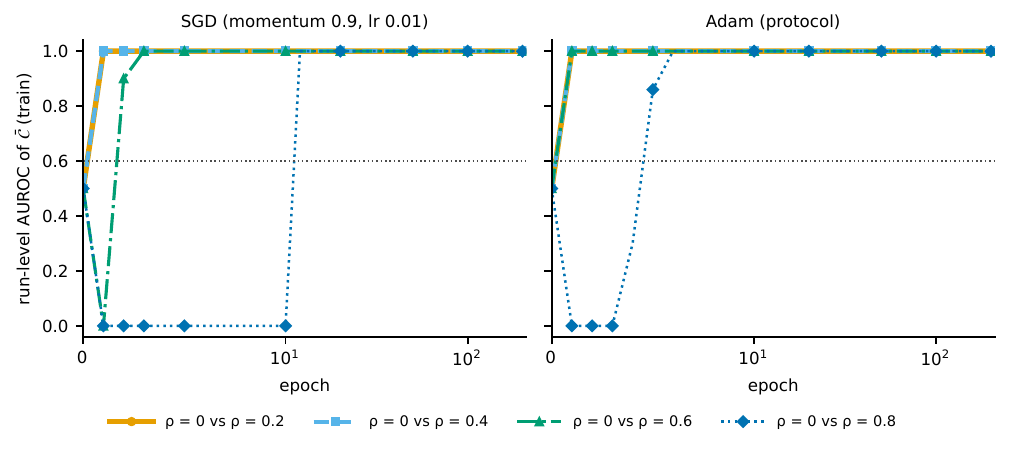}
\end{center}
\caption{\textit{Discrimination after interpolation}. Run-level AUROC of training-set $\Cb$ for distinguishing runs with $\rho=0$ from runs with $\rho>0$, by epoch. Left: SGD, for which all runs reached interpolation and clean runs did so at a median epoch of 18 and noisy runs at median epochs of 113-122. Right: the Adam protocol, under which no corrupted run reached interpolation. The dotted line indicates the pre-specified threshold of 0.6.}
\label{fig:h7}
\end{figure}

\paragraph{After interpolation (H7, post hoc).} H7 predicted that, once the network had interpolated and its training predictions had become sharp, $\Cb$ would no longer distinguish noisy from clean runs. Here, discrimination is measured by the AUROC between runs with $\rho=0$ and runs with $\rho>0$, using the run-level $\Cb$ computed on the training set. With SGD, the AUROC was $1.00$ both at the interpolation epoch and at epoch 200 for each noise rate considered (Figure~\ref{fig:h7}). At epoch 200, $\Cb$ ranged from $3.9\times10^{-3}$ at $\rho=0.2$ to $8.0\times10^{-3}$ at $\rho=0.8$, compared with $4.0\times10^{-4}$ for clean labels. This comparison is confounded by training time: the median interpolation epoch was $18$ for clean SGD runs, whereas it ranged from $113$ to $122$ for
noisy runs.

When runs were compared at matched times after interpolation, between 20 and 80 epochs, the ratio of $\Cb$ in noisy runs to that in clean runs ranged from $2.4$ to $4.3$ at $\rho=0.2$ and from $5.3$ to $9.2$ at $\rho=0.8$. The corresponding ratios at equal training epochs were $9.9$ and $20$. The AUROC remained $1.00$ throughout the matched-time comparison, although only 4 runs at $\rho=0.2$ and 6 runs at $\rho=0.4$ reached 80 epochs after interpolation. Matching runs by training loss on the observed labels, at loss levels of $0.03$, $0.01$, or $0.003$, further reduced the ratio to values between $1.2$ and $1.8$, while the AUROC remained between $0.97$ and $1.00$. These loss levels were reached at substantially different training times: between epochs $11$ and $30$ in clean runs and between epochs $88$ and $200$ in noisy runs. No noisy run reached a loss of $0.001$. Thus, neither comparison fully removes the effect of training time. Corrupted samples retained complexity values between $8\times10^{-3}$ and $9\times10^{-3}$ at every noise rate. Clean samples within noisy runs also retained higher complexity than samples in clean runs, with values of $2.6\times10^{-3}$ at $\rho=0.2$ and $7.3\times10^{-3}$ at $\rho=0.8$. Within the training horizon considered, $\Cb$ therefore continued to separate noisy from clean runs after interpolation, although training time could not be fully matched. After interpolation, $\Cb$ decreased in both groups, with a proportionally faster decline in clean runs. Consequently, the absolute difference narrowed while the ratio increased. Whether the difference vanishes with longer training remains open. Under the pre-specified criterion of a final AUROC $\leq0.6$, this post hoc analysis did not support H7.

\begin{table}[t]
\caption{\textit{Robustness summary}. Exploratory analyses, with 10 seeds per rate. Spearman ($r_s$): correlation of the event time with $\rho$ over $0<\rho<1$. Early stopping values are mean losses  in clean test accuracy, in percentage points, over $0<\rho<1$, evaluated on the evaluation half of the test set. Interpolation is the fraction of corrupted runs reaching 99.9\% training accuracy against noisy labels.}
\label{tab:robust}
\begin{center}
\small
\setlength{\tabcolsep}{3.5pt}
\begin{tabular}{p{0.12\linewidth}cccccccc}
\multicolumn{1}{c}{\bf }  &\multicolumn{1}{c}{\bf }  &\multicolumn{1}{c}{\bf }  &\multicolumn{2}{c}{\bf SPEARMAN $r_s$}  &\multicolumn{1}{c}{\bf }  &\multicolumn{1}{c}{\bf }  &\multicolumn{1}{c}{\bf }  &\multicolumn{1}{c}{\bf }
\\ \cline{4-5} \\
\multicolumn{1}{c}{\bf CONDITION}  &\multicolumn{1}{c}{\bf $\rho$}  &\multicolumn{1}{c}{\bf INTERP.}  &\multicolumn{1}{c}{\bf $t^*(\Cb)$}  &\multicolumn{1}{c}{\bf $t_{\rm best}$}  &\multicolumn{1}{c}{\bf FINAL}  &\multicolumn{1}{c}{\bf $\arg\max\Cb$}  &\multicolumn{1}{c}{\bf \begin{tabular}[c]{@{}c@{}}NOISY.\\ VAL. 1000\end{tabular}}  &\multicolumn{1}{c}{\bf \begin{tabular}[c]{@{}c@{}}LABEL\\WAVE $k1p3$\end{tabular}}
\\ \hline \\
MNIST, \textsf{UI}, Adam (protocol) & 0.2-0.8 & 0.00 & 0.93 & $-0.67$ & 40.3 & 22.9 & 0.8 & 0.4\\
MNIST, \textsf{UI}, Adam cosine & 0.2-0.8 & 0.75 & 0.93 & $-0.58$ & 38.3 & 23.0 & 0.7 & 0.3\\
MNIST, \textsf{UI}, Adam w.d.\ $10^{-4}$ & 0.2-0.8 & 0.03 & 0.94 & $-0.68$ & 39.5 & 23.6 & 0.8 & 0.3\\
MNIST, \textsf{UI}, SGD & 0.2-0.8 & 1.00 & 0.96 & $-0.82$ & 39.8 & 25.7 & 0.7 & 0.3\\
Fashion-MNIST, \textsf{UI}, Adam & 0.2-0.8 & 0.00 & 0.95 & $-0.69$ & 35.7 & 18.9 & 1.2 & 0.9\\
MNIST, PF, Adam & 0.1-0.4 & 0.03 & $-0.22$ & $-0.25$ & 18.2 & 4.7 & 0.7 & 0.9\\
Fashion-MNIST, PF, Adam & 0.2, 0.4 & 0.00 & 0.27 & $-0.75$ & 20.1 & 4.0 & 2.0 & 1.7\\
\hline
\end{tabular}
\end{center}
\end{table}

\section{Discussion}
\label{sec:discussion}

Table~\ref{tab:hypsum} relates the outcome of each hypothesis to the research questions.

\begin{table}[t]
\caption{\textit{Hypotheses, outcomes and interpretation}. Decisions follow the pre-specified criteria (Table~\ref{tab:hypdef}).}
\label{tab:hypsum}
\begin{center}
\small
\begin{tabular}{l l l p{0.62\linewidth}}
\multicolumn{1}{c}{\bf }  &\multicolumn{1}{c}{\bf RQ}  &\multicolumn{1}{c}{\bf DECISION}  &\multicolumn{1}{c}{\bf INTERPRETATION}
\\ \hline \\
H1 & -- & pass & Implementation check of exact identities.\\
H2 & RQ1 & pass & The memorization term behaved as expected on held-out and training data; on held-out data the null holds by construction, so this checks calibration.\\
H3 & RQ1 & pass & $\MI$ and $M$ agreed in ordering over epochs, a lenient criterion; their timings differed (Table~\ref{tab:mi_vs_m}).\\
H4 & RQ2 & fail & The peak of $V$ fell within a wide memorization window; that of $I(\Yh;X\mid\Yt)$ did not single out the phase.\\
H5 & RQ2 & pass & The peak of $\Cb$ beat lenient baselines, but not established rules at $\rho\ge0.6$.\\
H6 & RQ2 & fail & Per-sample complexity was weaker than label-based and label-free scores.\\
H7 & RQ2 & n/a & Not testable under the protocol; post hoc on SGD, $\Cb$ still separated noisy from clean runs.\\
H8 & -- & pass & Noise models were not interchangeable at complete corruption in our runs.\\
H9 & RQ2 & fail & The peak time depended on temperature more than on the functional.\\
H10 & RQ2 & fail & Seed-to-seed variation of centroid trajectories exceeded the separation between some rates.\\
\hline
\end{tabular}
\end{center}
\end{table}

\paragraph{What the memorization term measures (RQ1).} The budget in Proposition~\ref{prop:budget} is an accounting device built from standard identities. Several of our checks confirm the implementation rather than the measure: the identities (H1), the Fano bound and the end point hold by construction, and on held-out data the null holds by construction, where the permutation test rejected slightly more often than its nominal level. The empirical content is narrower. On the training set, $\MI$ rose together with $M$ (H2 and H3) and reached its complete-memorization value in runs that interpolated, while its held-out counterpart stayed near zero. 

Two qualifications are important. First, in the regimes studied, $\MI$ was close to what the training accuracies on clean and corrupted samples imply, and within a run it was close to a monotone function of $M$. Its practical advantage over the raw memorized fraction is mainly the absence of a chance level, which mattered most at $\rho=1$ and under pair-flip noise and which a chance-corrected $\kappa$ also largely removes. Second, $\MI$ measures a different notion from counterfactual memorization~\citep{FeldmanZhang2020,ZhangCounterfactual2023}: it is a property of the whole prediction map rather than the influence of a sample on its own prediction. Relating these notions and testing the null on held-out training samples rather than the test set are natural next steps. The term is a function-space counterpart of the label-noise information in the weights studied by \citet{Harutyunyan2020}, although we do not claim a formal relation. Because it requires clean labels, we used it as a reference quantity rather than a diagnostic.

\paragraph{Noise models (H8).} Under complete \textsf{UI} corruption, the training labels are complementary labels, and the network learned from them: test accuracy fell below chance while the predictions retained information about the true class, unlike under \textsf{SU} and \textsf{PERM}. Full \textsf{UI} corruption is therefore not a substitute for random labels in our setting.

\paragraph{What the mean complexity tracked (RQ2).} No pre-specified hypothesis addressed this question, and the following reading is exploratory. $\mathcal C(p)$ is non-monotone in the confidence of $p$, and the peak of $\Cb$ occurred when the confidences of the training samples passed through an intermediate range. At low noise this happened before memorization began. At high noise it happened during memorization, close to the epoch at which the mean confidence reached about 0.6, and our design cannot separate the rise in confidence from memorization there. On clean labels and under pair-flip noise, confidence rose within the first epochs and the peak occurred at epochs 1-2, which per-epoch logging cannot resolve. A model with homogeneous confidence did not predict the peak, which suggests that the spread of confidences matters. This reading is consistent with the shift in peak time with $\rho$ under \textsf{UI} noise, with its sensitivity to temperature (H9), and with stopping at the peak working at low noise and failing at high noise (H5 and the comparison with established rules). The pre-specified expectation that $V$ and the label-free heterogeneity term would mark the memorization phase (H4) was supported only for $V$, and only in a weak sense, because the phase spans many epochs. After  interpolation, $\Cb$ still distinguished noisy from clean runs within the observed horizon, contrary to H7, also at matched training loss although by a smaller factor, possibly through residual uncertainty on corrupted samples. The centroid trajectories in the entropy-disequilibrium plane were not stable across seeds relative to the separation between corruption rates (H10) and did not reveal the heterogeneity of the predictions (Appendix~\ref{app:geometry}). In our runs, the plane added little beyond the per-sample summaries. As decision tools, the established methods included in our comparisons performed better at high noise (H5 and H6).

\paragraph{Limitations.} The study used one architecture, a multilayer perceptron, and two small image datasets. It considered only synthetic class-conditional noise (Assumption~\ref{ass:A1}). Instance-dependent noise violates the assumption underlying Proposition~\ref{prop:null}, in which case the memorization term may also capture learnable patterns in the noise. Under the pre-specified protocol (Adam with a constant learning rate), no corrupted run reached interpolation. Interpolation and H7 could therefore be studied only post hoc, on SGD and cosine-schedule runs, with approximately 80 epochs of training after interpolation. The extension was defined after the main results and is exploratory. On Fashion-MNIST at $\rho=0.8$, $M$ reached 0.9 in only one of 10 runs, so $t_{90}$ is censored in that condition.  Furthermore, Label Wave was re-implemented from its description rather than the authors' code; the conclusion that the peak of $\Cb$ was outperformed at high noise also holds against noisy validation, which does not depend on this re-implementation. The selection of a label-free rule used clean labels on the validation half, and selection with noisy validation labels was not evaluated. Progressive early stopping~\citep{Bai2021} and counterfactual memorization were not evaluated. The hypotheses were fixed in an internal protocol, not in a public registry, and some criteria turned out to be lenient (H3, H4 and H5) or very hard to meet (H6 and H10). In the main grid the noise seed was tied to the initialization seed. Finally, plug-in information estimates are biased in finite samples; we addressed this using permutation nulls (Section~\ref{sec:valid}).


\section{Conclusion}
We asked how much information about label noise a classifier's predictions reflect and how it relates to accuracy-based measures of memorization (RQ1), and what statistical complexity tracks during training and whether it is useful as a tool (RQ2). Of the ten pre-specified hypotheses, five were supported, four were not, and one could not be tested under the protocol.

For RQ1, standard information identities decompose the information in a classifier's predictions into signal, memorization and heterogeneity terms. In our experiments, the in-sample memorization term rose with the memorized fraction while its held-out counterpart stayed near zero, but it was close to what the training accuracies already implied; its main advantage over the raw memorized fraction, the absence of a chance level, could largely be recovered through chance correction. Under complete corruption with incorrect labels, the network learned from the complementary labels and reached below-chance test accuracy, so in our runs this form of corruption was not equivalent to random labels.

For RQ2, the peak of the mean statistical complexity preceded memorization at low noise and occurred during memorization at high noise, close to the epoch at which the mean training confidence reached about 0.6; confidence and memorization could not be separated there. In our comparisons, mean complexity was not competitive as a stopping rule at high noise, nor as a detector of corrupted samples. An exploratory extension to Fashion-MNIST, pair-flip noise and four training variants gave similar results under uniform noise, although under pair-flip noise the complexity peak occurred within the first two epochs. These conclusions are limited to one architecture, two small datasets and synthetic noise; their applicability to convolutional architectures, larger datasets and real label noise remains to be tested.





\bibliography{reference}
\bibliographystyle{tmlr}

\appendix
\section{Geometry of the entropy-disequilibrium plane}
\label{app:geometry}
 
The boundaries of the region attainable by single distributions in the $(S,\mathcal D)$ plane, and the maxima of $\mathcal{C}=S\cdot\mathcal{D}$, were computed numerically for $N=10$ over three parametric families of candidate extremal distributions: two-level, ``partial'' and ``flat-at-$1/N$''. For $\mathcal D_2$ these families contain the extremal distributions of the entropy-index-of-coincidence diagram~\citep{HarremoesTopsoe2001}, so the boundaries are exact up to the resolution of the grid; for $\mathcal D_1$ and $\mathcal D_{JS}$ they are numerical estimates. As a check, we drew 120\,000 Dirichlet samples (20\,000 for each concentration parameter in $\{0.02, 0.1, 0.3, 1, 3, 30\}$). For every functional, none fell outside the numerical boundaries by more than $2\times10^{-3}$, and none exceeded the numerical maximum of $\mathcal C$. This check is necessary rather than sufficient, because random samples need not reach the extremal distributions.
 
\begin{center}
\small
\begin{tabular}{lccc}
\multicolumn{1}{c}{\bf FUNCTIONAL} & \multicolumn{1}{c}{\bf NORMALIZATION} & \multicolumn{1}{c}{\bf MAXIMUM OF $\mathcal{C}$} & \multicolumn{1}{c}{\bf MAXIMIZER: FAMILY,}\\
 & & \textbf{($N=10$)} & \textbf{LARGEST $p_k$}
\\ \hline \\
$\mathcal{D}_2$ & $(N-1)/N$ & 0.2519 & two-level, 0.756\\
$\mathcal{D}_1$ & $2(1-1/N)$ & 0.4388 & partial, 0.291\\
$\mathcal{D}_{JS}$ & value at a vertex & 0.3143 & partial, 0.250\\
\hline
\end{tabular}
\end{center}
 
For $\mathcal{D}_{JS}$, the Jensen-Shannon divergence between a vertex and the uniform distribution is $0.526$ nats, that is, approximately $0.758\,\ln 2$ for $N=10$. Normalizing by $\ln 2$ would therefore not map $\mathcal D_{JS}$ onto $[0,1]$, which is why we normalize by its value at a vertex.
 
\begin{figure}[h]
\begin{center}
\includegraphics[width=0.55\textwidth]{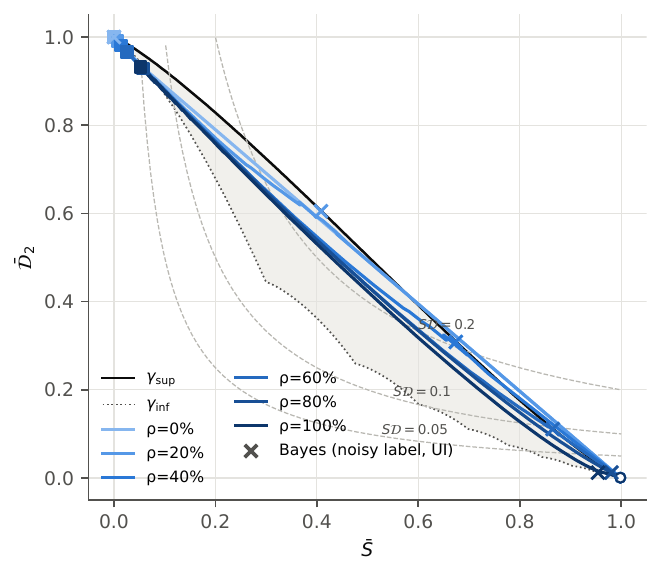}
\end{center}
\caption{\textit{Centroid trajectories in the entropy-disequilibrium plane}. Trajectories of $(\Sb(t),\Db_2(t))$ under \textsf{UI} noise, averaged over seeds, from epoch 0 (open circle) to epoch 200 (square). The shaded area is the region attainable by single distributions for $N=10$, bounded above by $\gamma_{\sup}$ (solid) and below by $\gamma_{\inf}$ (dotted). Crosses mark, for each $\rho$, the point of the Bayes posterior of the noisy label under \textsf{UI} noise, which assigns $1-\rho$ to the clean label and $\rho/9$ to each other class. A network predicting this posterior for every sample would reach it. Dashed lines are level sets of $S\cdot\mathcal{D}_2$ ($0.05$, $0.1$, $0.2$), not of $\Cb$.}
\label{fig:plane}
\end{figure}

\begin{figure}[h]
\begin{center}
\includegraphics[width=0.8\textwidth]{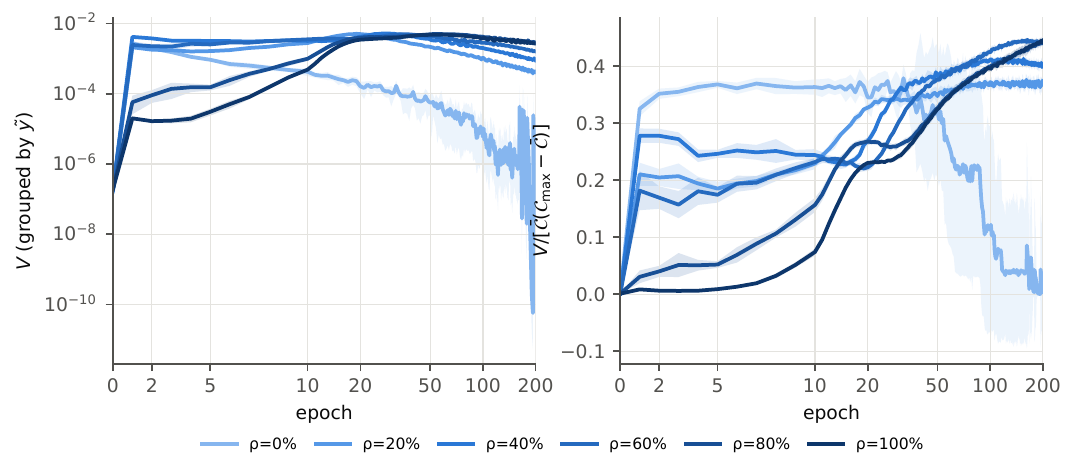}
\end{center}
\caption{\textit{Within-group variance}. Left: within-group variance $V$, with groups defined by the noisy label (logarithmic scale). Right: $V$ normalized by the bound of Observation~\ref{obs:mixture}, with $\mathcal C_{\max}=0.2519$ for $\mathcal D_2$ (Appendix~\ref{app:geometry}). \textsf{UI} noise, mean $\pm$ standard deviation over 10 seeds (Section~\ref{sec:mechanism}).}
\label{fig:variance}
\end{figure}

Let $\gamma_{\sup}(s)$ denote the numerical upper boundary of the region, that is, the largest $\mathcal{D}_2$ attainable by a single distribution with normalized entropy $s$. Because $\Sb$ and $\Db_2$ are averages, the centroid $(\Sb,\Db_2)$ always lies in the convex hull of the region, but it may lie above $\gamma_{\sup}$; in that case no single distribution has the same mean entropy and mean disequilibrium, and the difference $\Db_2 - \gamma_{\sup}(\Sb)$ would certify that the cloud is heterogeneous. In the main grid, this difference was positive in only 3 of the 16\,080 run-epoch pairs (\textsf{UI}, $\rho=0.6$, epochs between 1 and 2), with a largest value of $3.4\times10^{-4}$. This is below the resolution of the numerical boundary, so none of these epochs provides a certificate. The centroid trajectories therefore stayed within the region (Figure~\ref{fig:plane}), and the centroid alone did not reveal the heterogeneity of the predictions, which is visible in the budget (Section~\ref{sec:valid}) and in $V$ (Figure~\ref{fig:variance}).
 
\section{Label Wave configurations}
\label{app:labelwave}
 
Label Wave~\citep{YuanLabelWave2024} monitors the fraction of training samples whose predicted class changes between consecutive evaluations, smooths it with a trailing moving average over the last $k$ evaluations, and stops once the smoothed value has failed to improve on its running minimum for $p$ consecutive evaluations; the selected epoch is that of the minimum. In our re-implementation, the evaluation interval is one epoch and the predictions are computed on the full training set (Section~\ref{sec:protocol}). Table~\ref{tab:lw} reports the nine configurations of the grid $k\in\{1,3,5\}$, $p\in\{3,5,10\}$.
 
\begin{figure}[h]
\begin{center}
\includegraphics[width=\textwidth]{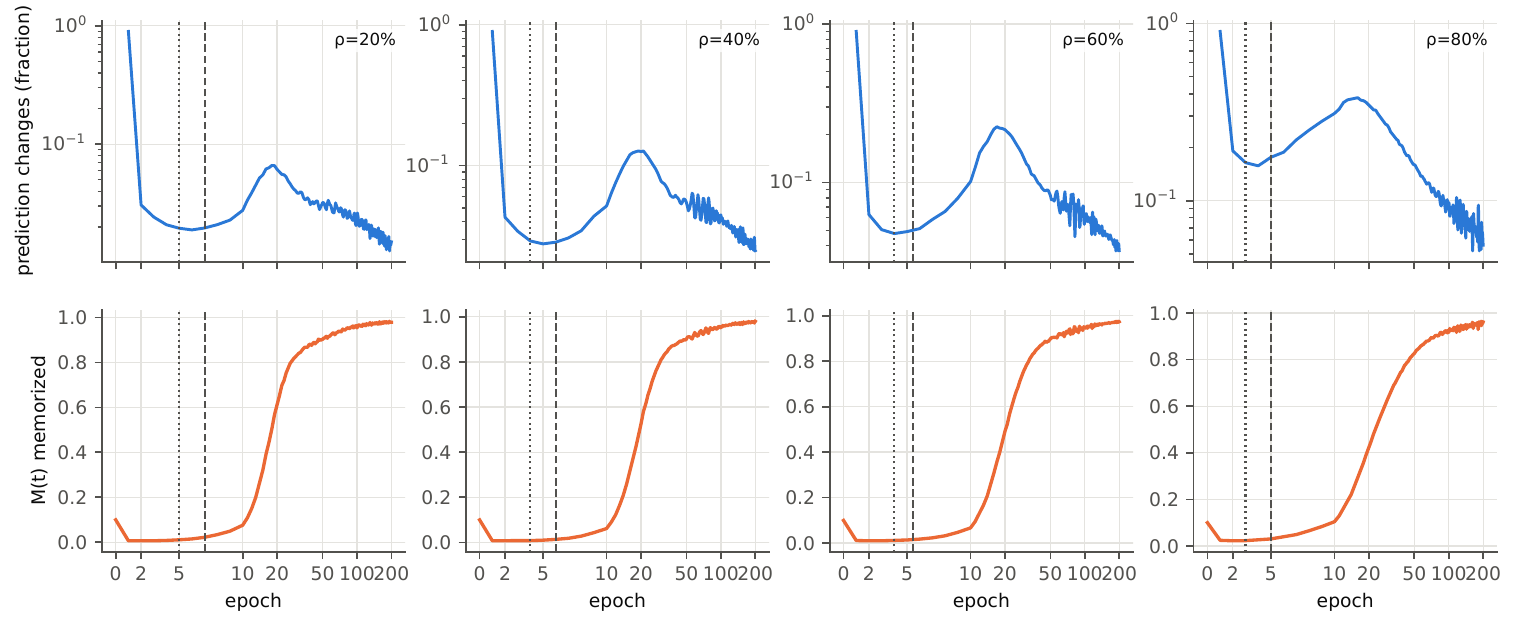}
\end{center}
\caption{\textit{Label Wave signal and memorized fraction}. Top: fraction of training samples whose predicted class changes between consecutive epochs (the Label Wave signal; logarithmic scale). Bottom: memorized fraction $M(t)$. Curves are means over 10 seeds under \textsf{UI} noise. Vertical lines mark the median over seeds of the Label Wave stopping epoch ($k=3,p=5$; dashed) and of the best epoch on the evaluation half of the test set (dotted) (Section~\ref{sec:tools}).}
\label{fig:lwsignal}
\end{figure}

For $0<\rho<1$, the stopping epoch did not depend on $p$ in any of the 40 runs. The prediction-change signal reached its minimum within the first epochs and then rose during memorization (Figure~\ref{fig:lwsignal}), so that every patience value stopped at the same minimum. The median stopping epochs were 4-6 for $k=1$, 5-7 for $k=3$ and 6-8 for $k=5$, against median best epochs of 3-5 on the evaluation half. A wider window thus delayed stopping by about one epoch per two additional epochs of smoothing. This delay mattered most at $\rho=0.8$, where test accuracy fell quickly after its maximum: the loss grew from 1.05 ($k=1$) to 3.03 ($k=3$) and 5.62 ($k=5$) percentage points.
 
At $\rho=0$ and $\rho=1$, where the signal has no rise caused by memorization, the patience changed the stopping epoch in 9 of 10 runs at $\rho=0$ and in all runs at $\rho=1$. For example, with $k=1$ the median stopping epoch was 12, 19 and 37 for $p=3$, 5 and 10 at $\rho=0$, and 45.5, 63.5 and 121 at $\rho=1$. The losses nevertheless stayed below 0.6 percentage points, because test accuracy changed little over these epochs at $\rho=0$ and stayed below chance throughout at $\rho=1$.

\begin{table}[t]
\caption{\textit{Results for the nine Label Wave configurations}. Loss in clean test accuracy relative to the best epoch, in percentage points, averaged over 10 seeds and evaluated on the evaluation half of the test set under \textsf{UI} noise. The rows $(1,3)$ and $(3,5)$ are those reported, rounded, in Table~\ref{tab:es}. For $0<\rho<1$, configurations with the same window $k$ gave identical losses for every patience $p$; the patience changed the result only at $\rho=0$ and $\rho=1$.}
\label{tab:lw}
\begin{center}
\begin{tabular}{lcccccc}
\multicolumn{1}{c}{\bf $(k,p)$}  &\multicolumn{1}{c}{\bf $\rho=0$}  &\multicolumn{1}{c}{\bf $\rho=0.2$}  &\multicolumn{1}{c}{\bf $\rho=0.4$}  &\multicolumn{1}{c}{\bf $\rho=0.6$}  &\multicolumn{1}{c}{\bf $\rho=0.8$}  &\multicolumn{1}{c}{\bf $\rho=1$}
\\ \hline \\
(1, 3) & 0.53 & 0.22 & 0.14 & 0.15 & 1.05 & 0.34\\
(1, 5) & 0.41 & 0.22 & 0.14 & 0.15 & 1.05 & 0.28\\
(1, 10) & 0.20 & 0.22 & 0.14 & 0.15 & 1.05 & 0.20\\
(3, 3) & 0.58 & 0.41 & 0.33 & 0.78 & 3.03 & 0.30\\
(3, 5) & 0.54 & 0.41 & 0.33 & 0.78 & 3.03 & 0.28\\
(3, 10) & 0.34 & 0.41 & 0.33 & 0.78 & 3.03 & 0.22\\
(5, 3) & 0.52 & 0.57 & 0.62 & 0.94 & 5.62 & 0.30\\
(5, 5) & 0.48 & 0.57 & 0.62 & 0.94 & 5.62 & 0.30\\
(5, 10) & 0.36 & 0.57 & 0.62 & 0.94 & 5.62 & 0.24\\
\hline
\end{tabular}
\end{center}
\end{table}
 
\section{Additional figures}
 
\begin{figure}[h]
\begin{center}
\includegraphics[width=\textwidth]{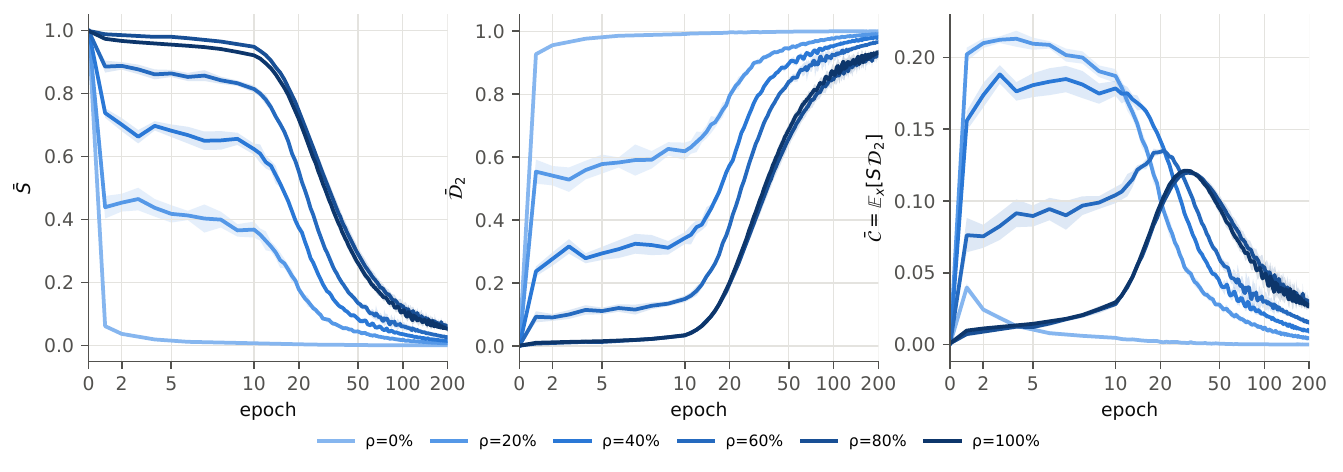}
\end{center}
\caption{\textit{Components of statistical complexity}. $\Sb$, $\Db_2$ and $\Cb$ on the training set under \textsf{UI} noise, mean $\pm$ standard deviation over 10 seeds (Section~\ref{sec:mechanism}).}
\label{fig:components}
\end{figure}
 
\begin{figure}[h]
\begin{center}
\includegraphics[width=\textwidth]{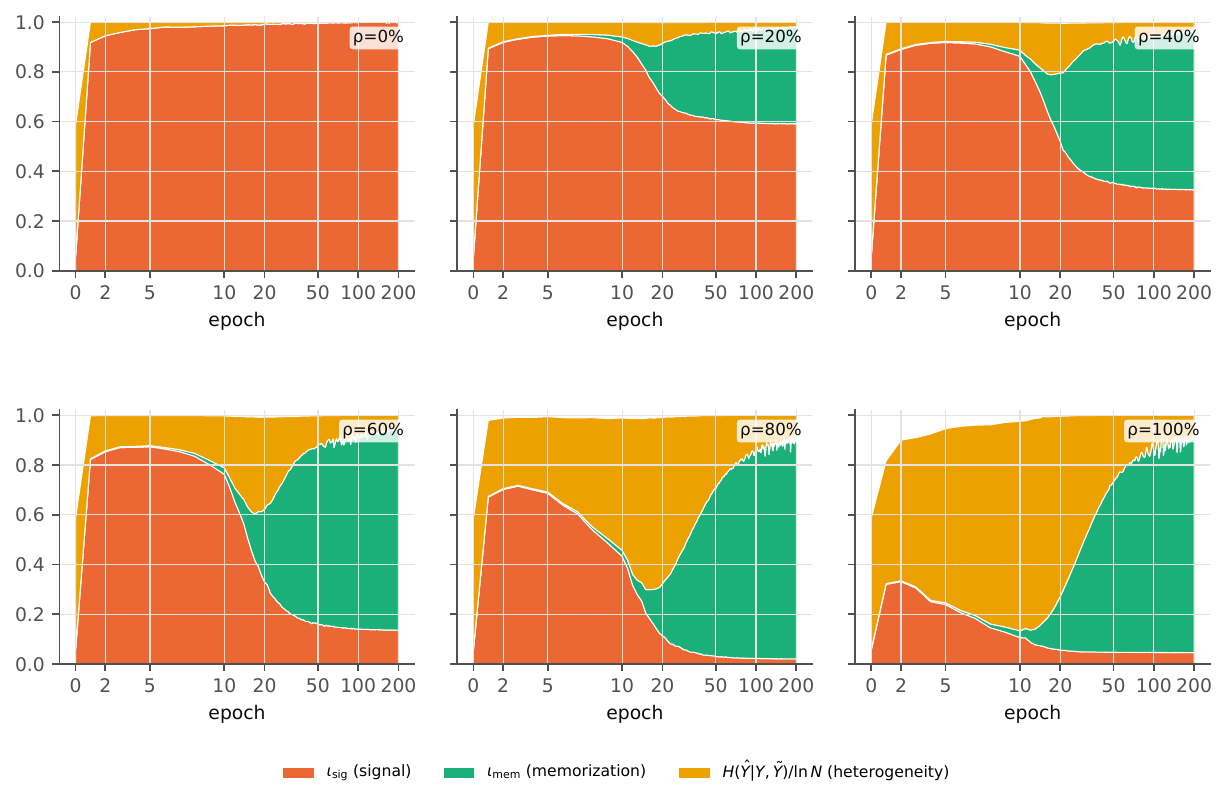}
\end{center}
\caption{\textit{Hard information budget}. Information budget computed from argmax predictions under \textsf{UI} noise, averaged over 10 seeds (Section~\ref{sec:valid}). Because argmax predictions are a deterministic function of the input, the mean conditional entropy is zero, the heterogeneity term equals $H(\Yh\mid Y,\Yt)/\ln N$, and the three terms add up to the normalized entropy of the predicted labels. These quantities do not change under temperature scaling.}
\label{fig:hardbudget}
\end{figure}
 
\begin{figure}[h]
\begin{center}
\includegraphics[width=0.48\textwidth]{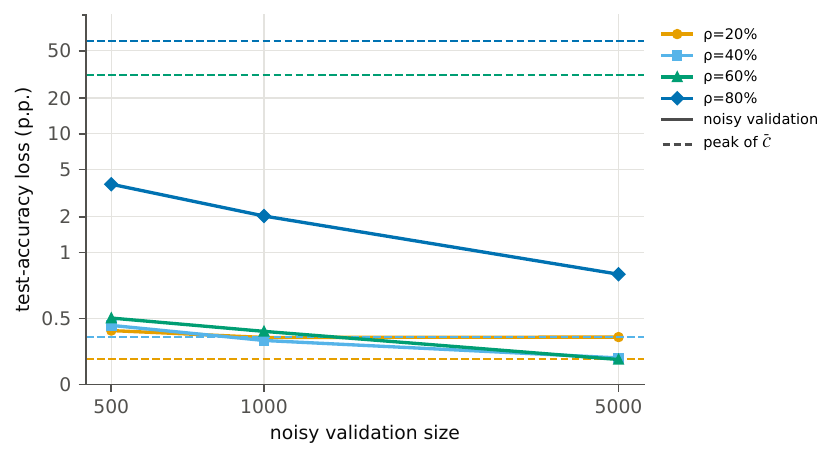}\hfill
\includegraphics[width=0.5\textwidth]{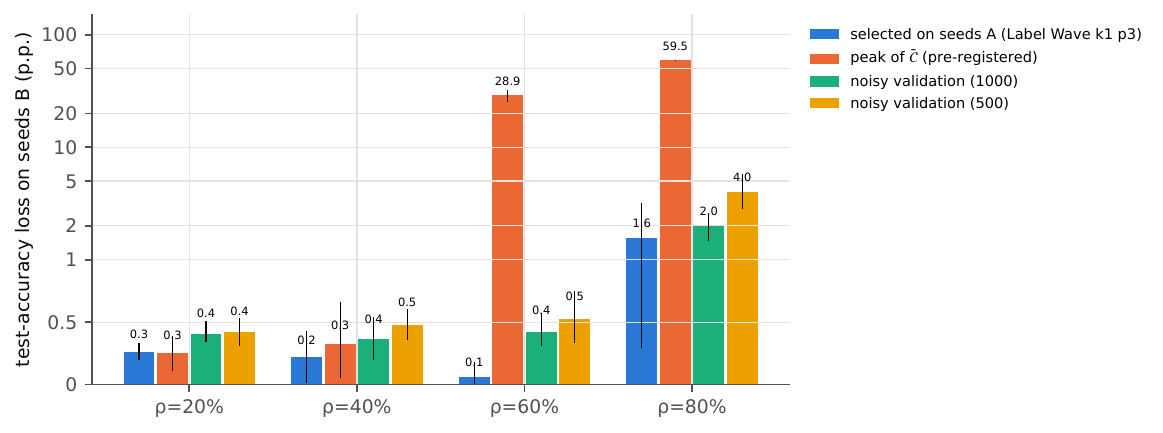}
\end{center}
\caption{\textit{Noisy-validation size and rule selection} (\textsf{UI} noise). Left: loss in clean test accuracy of the noisy-validation rule as a function of the validation-set size (solid lines and markers), and of stopping at the peak of $\Cb$ (dashed line of the same color), for each $\rho$; means over 10 seeds. Right: loss of the rule selected on five seeds and evaluated on the other five, compared with the peak of $\Cb$ and with noisy validation; bars are means with 95\% bootstrap intervals, and the numbers above the bars are the means. In this panel, selection and evaluation use the same test images; the selection on separate halves of the test set, reported in Section~\ref{sec:tools}, identified the same rule. Both panels use a symmetric-logarithmic vertical axis.}
\label{fig:valsize}
\end{figure}
 
\end{document}